\documentclass[sigconf]{acmart}

\usepackage{multirow}
\usepackage{hyperref}
\usepackage{xr}
\newcommand{\figref}[1]{Fig.~\ref{#1}}
\newcommand{\tabref}[1]{Table~\ref{#1}}
\newcommand{\equref}[1]{Eq.~\ref{#1}}
\newcommand{\secref}[1]{Sec.~\ref{#1}}

\def\nonneg#1{\tilde{#1}^{+}}
\def\negindex#1{{#1}^{-}}

\def\st{\rm st}
\def\ed{\rm ed}
\def\vr{\rm vr}
\def\tl{\rm tl}
\def\vcmr{\rm vcmr}
\def\threshold{\theta}

\def\eg{\emph{e.g.}} 
\def\ie{\emph{i.e.}} 
\def\etal{\emph{et al.}}

\AtBeginDocument{%
  \providecommand\BibTeX{{%
    \normalfont B\kern-0.5em{\scshape i\kern-0.25em b}\kern-0.8em\TeX}}}

\copyrightyear{2021}
\acmYear{2021}
\acmDOI{XX.XXXX/XXXXXXX.XXXXXXX}

\acmConference[arXiv '21]{arxiv '21}{2021}{arXiv}
\acmBooktitle{arXiv '21}
\acmPrice{0}
\acmISBN{XXX-X-XXXX-XXXX-X/XX/XX}
\begin{document}

\title{Video Moment Retrieval with Text Query Considering Many-to-Many Correspondence Using Potentially Relevant Pair}

\author{Sho Maeoki}
\affiliation{%
  \institution{The University of Tokyo}
  \city{Tokyo}
  \country{Japan}
}
\email{maeoki@mi.t.u-tokyo.ac.jp}

\author{Yusuke Mukuta}
\affiliation{%
  \institution{The University of Tokyo / RIKEN}
  \city{Tokyo}
  \country{Japan}
}
\email{mukuta@mi.t.u-tokyo.ac.jp}

\author{Tatsuya Harada}
\affiliation{%
  \institution{The University of Tokyo / RIKEN}
  \city{Tokyo}
  \country{Japan}
}
\email{harada@mi.t.u-tokyo.ac.jp}

\renewcommand{\shortauthors}{Maeoki, et al.}

\begin{abstract}
  In this paper we undertake the task of text-based video moment retrieval from a corpus of videos.
  To train the model, text-moment paired datasets were used to learn the correct correspondences.
  In typical training methods, ground-truth text-moment pairs are used as positive pairs,
  whereas other pairs are regarded as negative pairs.  
  However, aside from the ground-truth pairs, some text-moment pairs should be regarded as positive.
  In this case, one text annotation can be positive for many video moments.
  Conversely, one video moment can be corresponded to many text annotations.
  Thus, there are many-to-many correspondences between the text annotations and video moments.
  Based on these correspondences, we can form potentially relevant pairs,
  which are not given as ground truth yet are not negative;
  effectively incorporating such relevant pairs into training can improve the retrieval performance.  
  The text query should describe what is happening in a video moment.
  Hence, different video moments annotated with similar texts, which contain a similar action,   
  are likely to hold the similar action,
  thus these pairs can be considered as potentially relevant pairs.    
  In this paper, we propose a novel training method that takes advantage of potentially relevant pairs,
  which are detected based on linguistic analysis about text annotation.    
  Experiments on two benchmark datasets revealed that our method improves the retrieval performance both quantitatively and qualitatively.
\end{abstract}



\maketitle

\section{Introduction}
\label{sec:introduction}
\begin{figure}[t]
  \begin{center}    
   \includegraphics[width=1.0\linewidth]{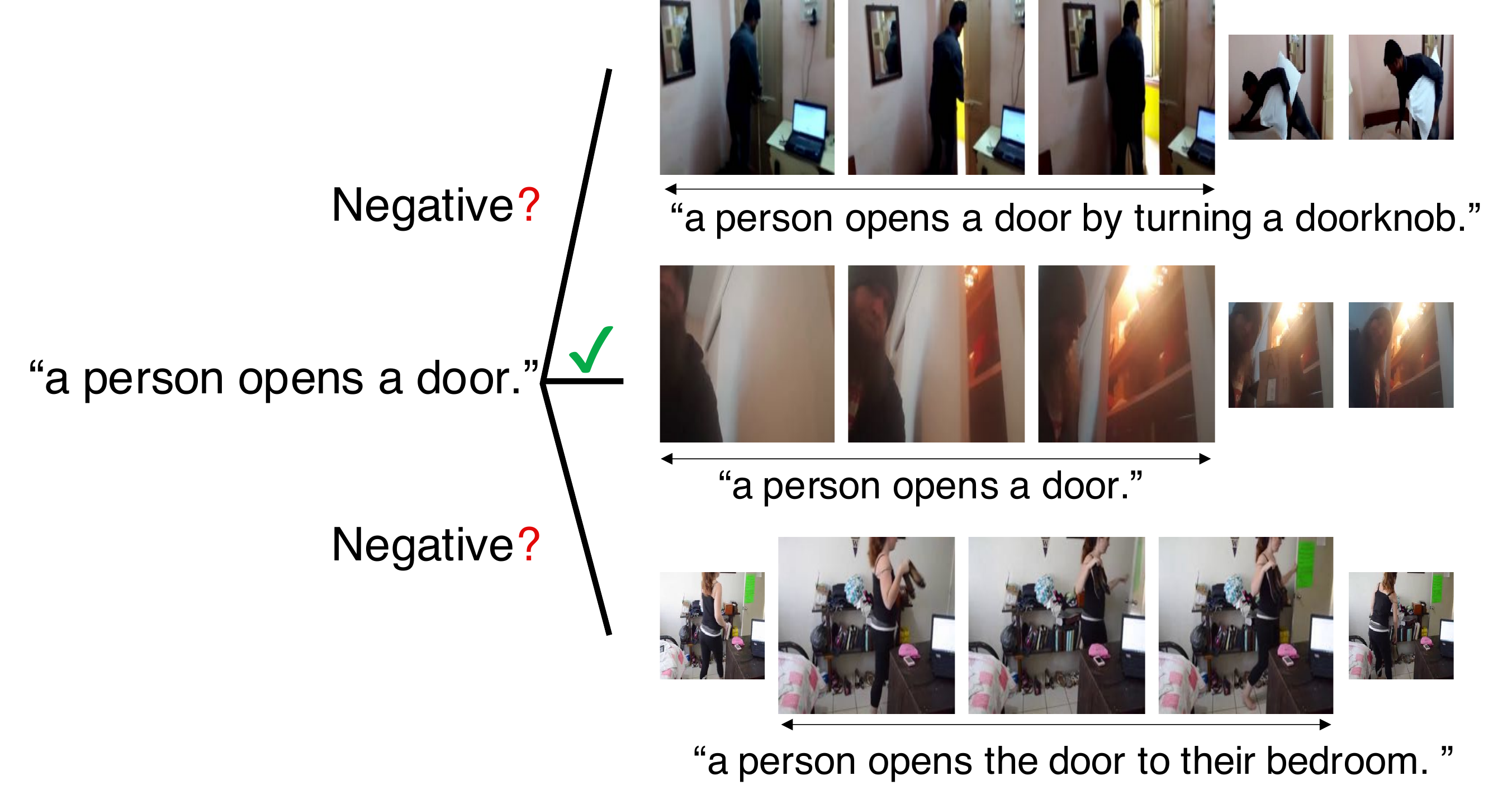}            
  \caption{Example of pairs that should not be treated as negative. 
  At the time of inference, all of the segments in this figure are said to be relevant to the query.}
  \label{fig:introduction_about_training}
  \end{center}
\end{figure}

In this study, we tackle the task of video moment retrieval from a corpus of videos via a text query~\cite{escorcia2019temporal}.
In this task, based on a text query that describes what a user wants,
the goal is to output the relevant moment from the video corpus.
In general, to train a model for text-based video retrieval, 
we use text-moment pairs offered in datasets.
While the ground-truth pairs, given in the datasets, are used as positive pairs,
other pairs are incorporated as negative pairs into training.
The effective use of the negative pairs can improve the model performance~\cite{faghri2018vse++}.
However, nonpositive pairs
are not necessarily negative.
For instance, as shown in \figref{fig:introduction_about_training},
we can see that there are relevant pairs expressing ``{\it a person opens a door}'' other than the ground-truth pairs,
any of which should not be interpreted as negative.
Rather, these pairs should be regarded as positive at the time of inference.
We call such relevant pairs that are not positive (ground-truth) but cannot be said negative ``\textit{potentially relevant pairs}.''.

\begin{figure}
  \begin{center}    
   \includegraphics[width=1.0\linewidth]{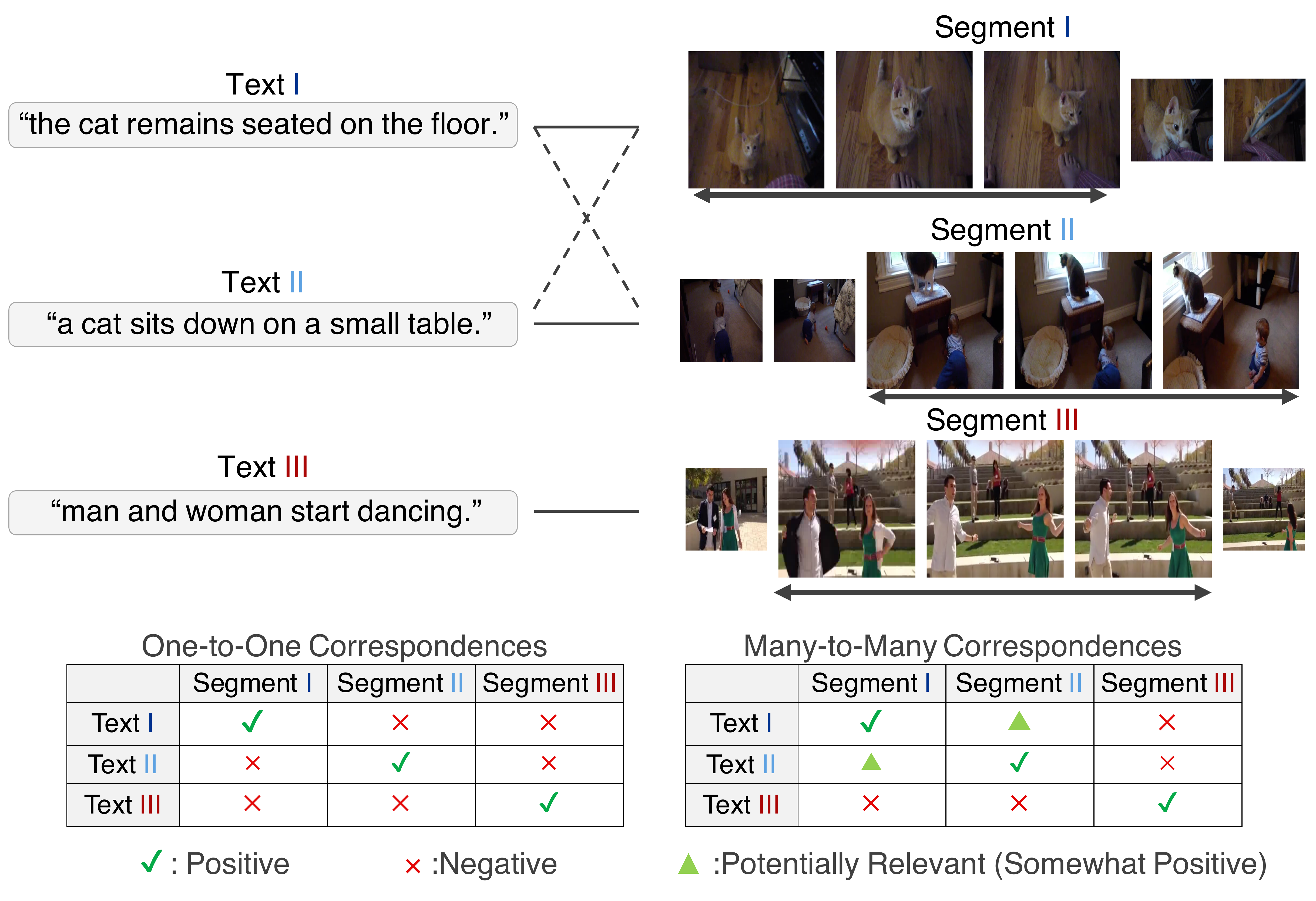}        
  \caption{Example of the many-to-many correspondence in practice.
  Though prior work utilizes the one-to-one correspondences of positive pairs,
  we attempt to incorporate the many-to-many correspondence of potentially relevant pairs as well as positive pairs into training.}
  \label{fig:method_focus}
  \end{center}
\end{figure}

If we can effectively utilize the potentially relevant pairs in training,
we can enhance the model performance consequently.
However, to utilize them, we need to determine the potentially relevant pairs,
which is non-trivial.
In practice, the potentially relevant pairs that contain almost the same content, as in \figref{fig:introduction_about_training}, are scarcely found in the dataset.
More generally, we assume that the potentially relevant pairs contain common elements, specifically similar actions,
and then attempt to detect the pairs based on linguistic analysis about text annotations.
For example, in \figref{fig:method_focus}, Segments I and II share the similar content, 
\ie, ``\textit{cat sits}''; both the Text I and Text II paired with the segments describe this element.
Here, we can form the many-to-many correspondence, as illustrated in \figref{fig:method_focus} on top of the ground-truth one-to-one correspondences,
and other pairs can be used as negative pairs because they do not seem to hold common elements.
Nevertheless, all potentially relevant pairs are not necessarily the same in terms of the pair validity (relevance).
Some are almost regarded as positive, whereas others are nearly negative, even though they might have common elements.
In fact, the potentially relevant pairs that can be formed in \figref{fig:introduction_about_training} are almost positive,
whereas the relevant pairs in \figref{fig:method_focus} are less relevant.

To address this problem, we incorporate a validity score into the loss functions by naming \textit{confidence}.
The loss functions include ranking loss and negative log-likelihood loss,
both of which are often used in this task~\cite{escorcia2019temporal, sudipta2020text,lei2020tvr,li2020hero,zhang2020hierarchical}.
In this way, we can handle the variance of the potentially relevant pairs.
We conducted experiments on two benchmark datasets (Charades-STA~\cite{gao2017tall} and DiDeMo~\cite{hendricks17iccv}) 
and revealed that using the approach on potentially relevant pairs can improve the model performance.
Our contributions are as follows:
\begin{itemize}    
    \item We proposed a novel training method to handle potentially relevant pairs as well as positive and negative pairs by extending loss functions.
    \item We proposed a method to detect potentially relevant pairs and measure their confidence scores (each pair relevance), based on linguistic analysis, specifically SentenceBERT.
    \item We showed training with potentially relevant pairs can improve the model performance quantitatively and qualitatively.    
\end{itemize}

\section{Related Work}
\subsection{Video Corpus Moment Retrieval}
\label{sec:vcmr_explanation}
Video corpus moment retrieval (VCMR) is a task that outputs relevant video moments from a corpus of videos, given a text query.
In terms of pipeline, most prior work on VCMR involved video-level retrieval and temporal localization (video moment retrieval).

\begin{figure*}
  \begin{center}
  \centering    
   \includegraphics[width=0.8\linewidth]{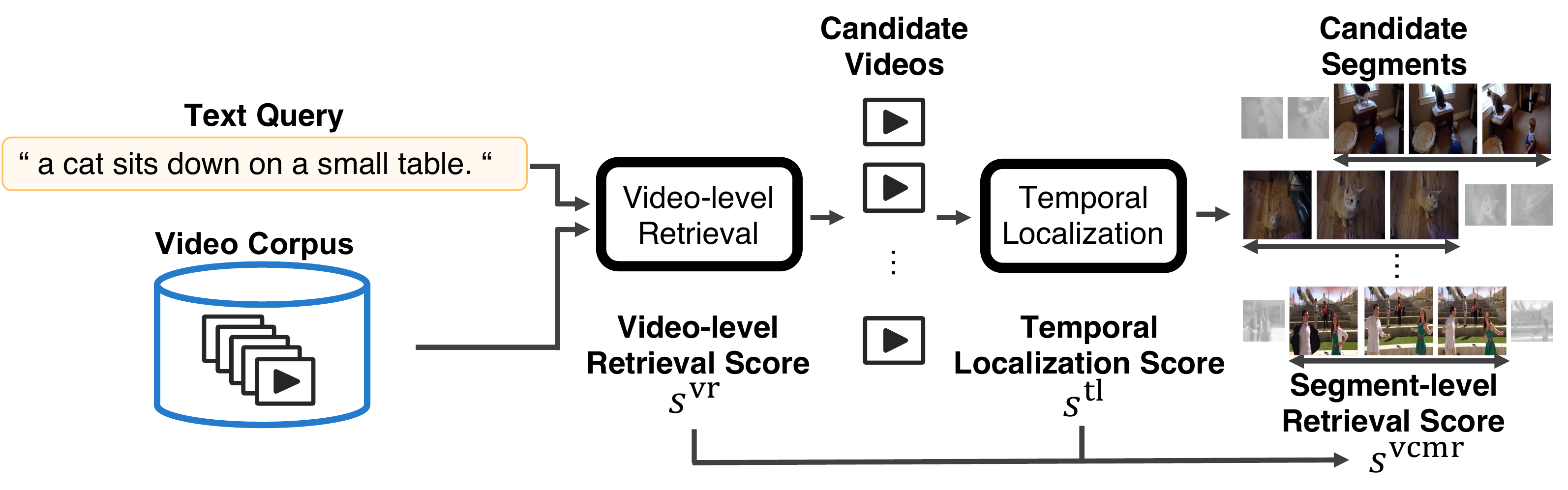}        
  \caption{Pipeline of video corpus moment retrieval. 
  A text query and a video corpus are fed into a model to compute the video-level retrieval score $s^{\vr}$;
  then candidate videos are selected based on this score.
  Next, the temporal localization score $s^{\tl}$ is computed for these candidate videos.
  The two scores are combined to compute the segment-level retrieval score $s^{\vcmr}$, and
  finally, the candidate segments are output based on this score.}
  \label{fig:baseline_pipeline_overview}
  \end{center}
\end{figure*}

    \textbf{Video-level retrieval (VR)} is a task to output an entire video that includes a video moment corresponding to a given query
    based on a predefined score, where higher is better.
   The model is typically trained using a loss function, such as ranking loss (\eg,~\cite{faghri2018vse++}),
   which increases the positive pair score while lowering the negative pair score~\cite{miech18learning, yu2018joint, mithun2018learning,Maeoki_2020_CVPR_Workshops, gabeur2020multi}.

    \textbf{Single video moment retrieval (SVMR)} was concurrently proposed by Gao~\etal~\cite{gao2017tall} and Hendricks~\etal~\cite{hendricks17iccv},
    enabling temporal localization of video moment.
    When a pair of text and a video are given,
    the model is required to predict both the start and end times, \ie, the most relevant moment in the video.
    A series of methods~\cite{liu2018tmn, hendricks18emnlp, zhang2019man, chen2018temporally} have been proposed based on \cite{gao2017tall,hendricks17iccv}.
    Unlike video corpus moment retrieval (VCMR), these methods basically assume that the input video paired with each text query is predetermined,
    which means SVMR need not detect relevant videos.

   \textbf{Video corpus moment retrieval (VCMR)} was first studied by Escorcia~\etal~\cite{escorcia2019temporal}, 
   in which they proposed a VCMR efficient method by combining VR and SVMR in their pipeline.
   Since Escorcia \etal, research on VCMR has progressed gradually~\cite{sudipta2020text,lei2020tvr,li2020hero,zhang2020hierarchical}.   
   In particular, Lei \etal~\cite{lei2020tvr} extended the VCMR to the text domain (subtitle) as well as the visual domain 
   by proposing a TVR dataset, targeting TV Show/Movie, and an even more efficient retrieval method (\ie, Cross-modal Moment Localization; XML).
   Despite its simple architecture, XML simultaneously achieves high performance and efficiency,
   hence we adopt it as our backbone.

\subsection{Training focusing on positive/negative relationships}
The potentially relevant pair assumes the incompleteness of the dataset labels,
as we assume the existence of positive pairs other than ground truth.
Kanehira \etal~\cite{kanehira2016multi} addressed the label incompleteness by applying positive and unlabeled (PU) theory~\cite{jaskie2019positive} to the multi-label classification problem~\cite{DBLP:conf/icml/DembczynskiCH10}.
Multi-label classification can be interpreted as a matching problem between the input (image) and predefined classes.
In this context, the text-based retrieval can be considered as an extension of the classes to text, which is free form and of countless patterns.
Our work attempts to identify potentially relevant pairs via linguistic analysis of text annotations.
This is equivalent to the explicit consideration of semantic relationships between classes,
which is more desirable, moreover could not be handled in \cite{kanehira2016multi}.
Our method is inspired mainly by two works~\cite{lin2020world,zhang-etal-2020-learning}.

Lin \etal~\cite{lin2020world} prepared external (\textit{grayscale}) data as well as 
randomly selected samples (negative) and ground-truth (positive) data to train a response selection system.
They incorporated the \textit{grayscale} data into a loss function \ie, ranking loss
in which the relationship between the three types of data is considered.
They termed this loss multi-level ranking loss and report that it can contribute to performance improvement.
In this work, we introduce a potentially relevant pair by incorporating it into the multilevel ranking loss.
However, unlike \cite{lin2020world}, obtaining the \textit{grayscale} data (potentially relevant pairs) is nontrivial because of the task difference.
Additionally, we need to handle the variance of such \textit{grayscale} data;
therefore, we incorporate the confidence score into the multilevel ranking loss.

Zhang \etal~\cite{zhang-etal-2020-learning} addressed cross-modal learning using text and images.
Similar to our research, they found common elements in the text and attempted to utilize them for learning.
They built a graph structure called a denotation graph~\cite{young-etal-2014-image} in advance,
then using the obtained graph, positive/negative samples were selected.
In contrast to Zhang \etal, we consider potentially relevant data, such as \textit{grayscale} data, in addition to positive/negative data.
Moreover, the localization problem in VCMR should be solved, which is not addressed in~\cite{zhang-etal-2020-learning}.
In addition to the task difference, because Zhang \etal~use the denotation graph, which is a method based on classical natural language processing,
similar contents with totally different words \eg, ``\textit{sit}'' and `` \textit{seated}'', are regarded as irrelevant.
To mitigate this problem, we adopt a neural-network-based soft metric to detect the potentially relevant pairs.

\section{Video Corpus Moment Retrieval}
\subsection{Task Overview}
Given a text query $q_{i}$ and a corpus of videos $\mathcal{V}$,
VCMR outputs the relevant moment starting at $t_{\st}^{i}$ and ending at $t_{\ed}^{i}$ 
in a video $v_{i} \in \mathcal{V}$, which we denote as $v_{i}[t_{\st}^{i}: t_{\ed}^{i}]$.
In VR, a model is required to output the video $v_{i}$ that contains the relevant moment.
On the other hand, in VCMR, a model needs to localize the start $t_{\st}^{i}$ and end time $t_{\ed}^{i}$ and,
consequently, output the video segment $v_{i}[t_{\st}^{i}: t_{\ed}^{i}]$.
To address this point, we estimate the timestamp as in prior work~\cite{lei2020tvr} because of its efficiency and good performance.

Meanwhile, in SVMR, the text query $q_{i}$ and its corresponding video $v_{i}$ are fed into a model,
then the model infers the relevant moment timestamp.
This is different from VCMR in that a model is required to detect which video contains the relevant moment,
\ie, the model needs to distinguish $v_{i}$ from other irrelevant videos in the video corpus $\mathcal{V}$.
This characteristic results in the following problem;
the text query $q_{i}$ paired with $v_{i}[t_{\st}^{i}: t_{\ed}^{i}]$ in the dataset 
can also correspond to another moment, $v_{j}[t_{\st}^{j}: t_{\ed}^{j}]$ paired with $q_{j}$.
Thus, the pair $(q_{i}, v_{j}[t_{\st}^{j}: t_{\ed}^{j}])$ can be regarded as a potentially relevant pair.
Likewise, the pair of $(q_{j}, v_{i}[t_{\st}^{i}: t_{\ed}^{i}])$ is also likely to be a potentially relevant pair.
In this case, we denote $j$ as $\nonneg{i}$.

Typically, nonpositive pairs are incorporated as negative pairs at the time of training~\cite{escorcia2019temporal, sudipta2020text,lei2020tvr,li2020hero,zhang2020hierarchical}.
However, utilizing such potentially relevant pairs as negative pairs would result in deteriorating the model performance.
Conversely, the retrieval performance can be improved by effectively using relevant pairs.
In this paper, we consider the many-to-many correspondences between Text I and Text II and Segment I and Segment II, as illustrated in \figref{fig:method_focus},
elaborating on loss design using the potentially relevant pair, through which semantic grounding of text to video can be properly learned.

\subsection{Backbone}
\paragraph{Architecture}
As described in \secref{sec:vcmr_explanation}, we adopt XML~\cite{lei2020tvr} as our backbone.
A text query $q_{i}$ composed of a set of words is encoded via the text encoder RoBERTa~\cite{liu2019roberta}.
Consequently, contextualized word-level embeddings $E^{q}_{i} = \{ \mathbf{e}_{i,n}^{q} \}_{n=1}^{l_{q}}$, where $ \mathbf{e}_{i,n}^{q} \in \mathbb{R}^{D_{q}}$, are obtained.
Note that $l_{q}$ represents the number of words in $q_{i}$, and $D_{q}$ is the word-embedding dimension.

On the other hand, a video $v_{j}$ is divided into a set of clips $v_{j} = \{ c_{j,n} \}_{n=1}^{l_v}$ every $\delta t$ second.
Each clip $c_{j,n}$ can be denoted as $c_{j,n}= v_{j}[(n-1)\delta t: n \delta t]$.
The clips were then converted, via a video encoder, into a set of clip-level video representations 
$E^{v}_{j} = \{ \mathbf{e}_{j,n}^{v} \}_{n=1}^{l_{v}}$, where $\mathbf{e}_{j,n}^{v} \in \mathbb{R}^{D_{v}}$.
The video encoder typically consists of convolutional neural networks including ResNet-152~\cite{he2016deep},
where frame-level features are pooled across the temporal direction to obtain the clip-level representations.
Note that $l_{v}$ represents the number of clips in $v_{j}$, and $D_{v}$ is the dimension of the clip-level video representation.

The contextualized word embeddings $E^{q}_{i}$ are transformed into a vector ${\bf q}_{i} \in \mathbb{R}^{d}$ via a transformer~\cite{vaswani2017attention},
whereas the clip-level video representations $E^{v}_{i}$ are converted into two sets of hidden vectors:
$H^{v,1}_{j} = \{ \mathbf{h}_{j,n}^{v,1} \}_{n=1}^{l_{v}}, H^{v,2}_{j} = \{ \mathbf{h}_{j,n}^{v,2} \}_{n=1}^{l_{v}}$,
where $\mathbf{h}_{j,n}^{v,1} \in \mathbb{R}^{d}, \mathbf{h}_{j,n}^{v,2} \in \mathbb{R}^{d}$
via another transformer.
Note that $d$ represents the dimension of the hidden vectors.
The obtained features ${\bf q}_{i}, H^{v,1}_{j},$ and $H^{v,2}_{j}$ are used to compute a score to achieve VCMR.

\paragraph{Inference}
\begin{equation}            
  s^{\vr}(v_{j}| q_{i}) = \max( S_\mathrm{cos}({\bf q}_{i}, H^{v,1}_{j})).
  \label{eq:vr_score}    
\end{equation}
\begin{equation}        
  s^{\tl}(\bar{t}_{\st}, \bar{t}_{\ed}| q_{i}, v_{j}) = P_{\st}(\bar{t}_{\st})P_{\ed}(\bar{t}_{\ed}).
  \label{eq:svmr_score}
\end{equation}
\begin{equation}    
    s^{\vcmr}(v_{j}, \bar{t}_{\st}, \bar{t}_{\ed}| q_{i}, \mathcal{V}) = 
        s^{\tl}(\bar{t}_{\st}, \bar{t}_{\ed}| q_{i}, v_{j}) \exp(\alpha s^{\vr}(v_{j}|q_{i})).
    \label{eq:vcmr_score}
\end{equation}

As illustrated in \figref{fig:baseline_pipeline_overview}, the retrieval is performed based on a score for VCMR $s^{\vcmr}$,
which is computed by combining a video-level retrieval score $s^{\vr}$ and temporal localization score $s^{\tl}$.
The score $s^{\vr}$ of $v_{j}$ w.r.t. $q_{i}$ is defined as \equref{eq:vr_score},
where $S_{\mathrm{cos}}$ outputs a similarity matrix composed of cosine similarities between all pairs of input sets.
Moreover, to infer starting time $t_{\st}$ and ending time $t_{\ed}$,
probability distributions $P_{\st} \in \mathbb{R}^{l_{v}}$, $P_{\ed} \in \mathbb{R}^{l_{v}}$ 
are obtained using ${\bf q}_{i}$ and $H^{v,2}_{j}$.
Then, the likelihoods w.r.t. $\bar{t}_{\st}, \bar{t}_{\ed}$,
which are denoted as $P_{\st}(\bar{t}_{\st}) \in \mathbb{R}, P_{\ed}(\bar{t}_{\ed}) \in \mathbb{R}$,
are used to define $s^{\tl}$ as in \equref{eq:svmr_score}.
Note that $\bar{t}_{\st} \leq \bar{t}_{\ed}$, furthermore
\begin{equation}
  \label{eq:inference_constraints}
  n_\mathrm{min} \delta t \leq \bar{t}_{\ed}-\bar{t}_{\st} \leq n_\mathrm{max} \delta t
\end{equation}
is given as a constraint.
When given a text query $q_{i}$ and a video corpus $\mathcal{V}$,
the VCMR score with respect to a video moment $v_{j}[\bar{t}_{\st}: \bar{t}_{\ed}]$ is summarized as \equref{eq:vcmr_score},
where $\alpha$ is a trade-off parameter between $s^{\vr}$ and $s^{\tl}$.
In practice, as per \cite{lei2020tvr}, the top 100 videos are selected based on $s^{\vr}$ first,
then $s^{\tl}$ is computed for the top 100 videos,
and finally, candidate segments are obtained according to $s^{\vcmr}$.

\section{Proposed Method}
\paragraph{Method Overview}
To consider the many-to-many correspondence mentioned in \figref{fig:method_focus},
we propose a training method which effectively uses potentially relevant pairs.
In this paper, we find out potentially relevant pairs among unlabeled pairs,
then distinguish them from negative pairs.
This process is performed via linguistic analysis using text annotations given in a dataset,
assuming described events in the text definitely happen in the corresponding moment.
However, owing to the variety of relevance in each pair, as discussed in \secref{sec:introduction},
we need to carefully incorporate the pairs into training, based on their relevance.
To achieve this, for each detected potentially relevant pair, 
we introduce a confidence score that indicates to what extent the pair is relevant.
The confidence score is also computed via linguistic analysis,
which is expected to give positive-like behavior for a higher value 
and negative-like behavior for a lower value.

\subsection{Potentially Relevant Pair}
\label{sec:detection_and_confidence}
If there is a common element between two text queries $q_{i}$ and $q_{j}$
in terms of semantics, notating $j=\nonneg{i}, i=\nonneg{j}$,
we call the pairs of ($q_{i}$, $v_{\nonneg{i}}[t_{\st}^{\nonneg{i}}: t_{\ed}^{\nonneg{i}}]$) 
and ($q_{\nonneg{j}}$, $v_{j}[t_{\st}^{j}: t_{\ed}^{j}]$)
potentially relevant pairs.
For these pairs, we compute the confidence scores $C_{(i,\nonneg{i})}, C_{(\nonneg{j}, j)}$ respectively.
These scores can be interpreted as \textit{confidence} that the given pairs are positive;
introducing such confidence score enables training that takes into account each pair relevance.
Because our target is video, the preferable common elements are actions (verbs) rather than objects (nouns).
In contrast, the confidence score $C$ should reflect overall semantic similarity.
This is because even if there is a common element between $q_{i}$ and $q_{j}$,
in the case that the overall semantics are totally different, 
what happens in $v_{i}[t_{\st}^{i}: t_{\ed}^{i}]$ is likely to be different enough from 
that of $v_{j}[t_{\st}^{j}: t_{\ed}^{j}]$.

In this study, we use text annotations for both the detection of potentially relevant pairs ($\Phi$) and confidence computation ($\Psi$).
However, the focus points of $\Phi$ and $\Psi$ are different.
For $\Phi$, we expect to capture local semantic common elements, 
whereas global semantic similarity should be considered for $\Psi$.
Below, the detection of common elements between $q_{i}$ and $q_{j}$ is expressed as 
\begin{equation}    
  f_{(i,j)} = \Phi(q_{i},q_{j}),
  \label{eq:non_negative_detection}
\end{equation}
where $f_{(i,j)}$ indicates zero/one when a given pair is regarded as negative/relevant.
The confidence score $C_{(i,j)}$ ($0 \leq C_{(i,j)} \leq 1$) is computed for the detected potentially relevant pair as per
\begin{equation}    
  C_{(i,j)} = \Psi(q_{i}, q_{j}|f_{(i,j)}=1).
  \label{eq:conf_calculation}
\end{equation}

\paragraph{Designing $\Phi$ and $\Psi$ using SentenceBERT}
Although we can design $\Phi$ and $\Psi$ using $n$-gram or its variants, represented by BLEU~\cite{papineni-etal-2002-bleu},
the $n$-gram may extract unexpected common elements, such as grammatical structure,
which is not necessarily related to what is happening in a particular moment.
To avoid this problem, we design $\Phi$ and $\Psi$ using SentenceBERT (SBERT)~\cite{reimers-gurevych-2019-sentence}.
SBERT is one of the pretrained neural networks for natural language processing,
obtained by finetuning BERT~\cite{devlin-etal-2019-bert} variants.
Cosine similarity between two SBERT embeddings gives higher/lower values
when the input two sentences are more/less relevant.
In designing $\Phi$ and $\Psi$ with SBERT, we need to consider the following two characteristics:
\begin{itemize}
  \item SBERT can measure overall semantic similarity between two sentences.
  \item Even if there is a common element, the output score becomes lower when the ratio of the element is small in the sentences.
 \end{itemize} 

The first characteristic suggests that SBERT seems suitable for $\Psi$.
However, the second characteristic implies that SBERT may not work for $\Phi$ if a sentence is lengthy or complicated.
To detect the potentially relevant pair,
we attempt to compare common elements such as actions in sentences in the form of short chunks.
Thus, we attempt to use verb phrases (VPs), rather than noun phrases (NPs), obtained by the constituency parser~\cite{Kitaev-2018-SelfAttentive},
then measure similarities between the elements such as VPs using SBERT.
Suppose that we denote a set of VPs and NPs obtained from the text $q_{i}$
as $\{ {\rm VP}^{i}_{n} \}_{n}$ and $\{ {\rm NP}^{i}_{n} \}_{n}$, respectively.
Then, $\{ {\rm VP}^{i}_{n} \}_{n}$ is expected to hold a set of actions described in the text.
For instance, suppose that we have a sentence ``\textit{man and woman start dancing}'' as in \figref{fig:method_focus},
we can extract ``\textit{man and woman}'' as a NP, ``\textit{start dancing}'' and ``\textit{dancing}'' as VPs.
As mentioned above, VPs contain sets of actions described in a given sentence.
Thus, we can determine whether there is a common element between $q_{i}$ and $q_{j}$
by comparing $\{ {\rm VP}^{i}_{n}  \}_{n}$ and $\{ {\rm VP}^{j}_{n}  \}_{n}$ and comparing $q_{i}$ and $q_{j}$ directly.

Let $\{ {\rm VP}^{i}_{n}  \}_{n}|_{n=0} = q_{i}, \{ {\rm VP}^{j}_{n}  \}_{n}|_{n=0}=q_{j}$.
Then, we compute the cosine similarity for all pairs of $\{ {\rm VP}^{i}_{n}  \}_{n}$ and $\{ {\rm VP}^{j}_{n}  \}_{n}$,
then, when any of the pairs is not less than a predefined threshold $\threshold$,
we judge this pair as a potentially relevant pair.
If we denote the SBERT embeddings of $\{ {\rm VP}^{i}_{n} \}_{n}$ as $\{ \bar{{\bf q}}^{\mathrm{vp}}_{i,n}  \}_{n}$,
the detection process can be formulated as 
\begin{equation}    
  \Phi_{\mathrm{vp}}(q_{i}, q_{j}) = \left\{
      \begin{array}{ll}
          1 & (\max(S_\mathrm{cos}(\{ \bar{{\bf q}}^{\mathrm{vp}}_{i,n}  \}_{n}, \{ \bar{{\bf q}}^{\mathrm{vp}}_{j,n}  \}_{n})) \geq \threshold), \\
          0 & (otherwise).
      \end{array}
      \right.
  \label{eq:vp_level_non_negative_detection}
\end{equation}

The confidence computation is performed by computing the cosine similarity between $q_{i}$ and $q_{j}$:
\begin{equation}    
  \Psi(q_{i}, q_{j}|f_{(i,j)}=1) = [S_\mathrm{cos}(\bar{{\bf q}}_{i}, \bar{{\bf q}}_{j})]_{+},
  \label{eq:conf_calculation_with_sbert}
\end{equation}
where $\bar{{\bf q}}_{i}$ and $\bar{{\bf q}}_{j}$ are the embeddings of $q_{i}$ and $q_{j}$, $[\cdot]_{+} = \max(\cdot, 0)$.

Instead of solely using VPs \eg, ``{\it start dancing}'' and ``{\it dancing}'' in addition to the original sentence ``\textit{man and woman start dancing},''
we also tried $\Phi_{\mathrm{np-vp}}$ that uses concatenation of words in NPs and VPs to form short sentences
\eg, ``{\it man start dancing}'' and ``{\it woman start dancing},'' with a simple rule-based algorithm.

\begin{figure}
  \begin{center}    
   \includegraphics[width=0.8\linewidth]{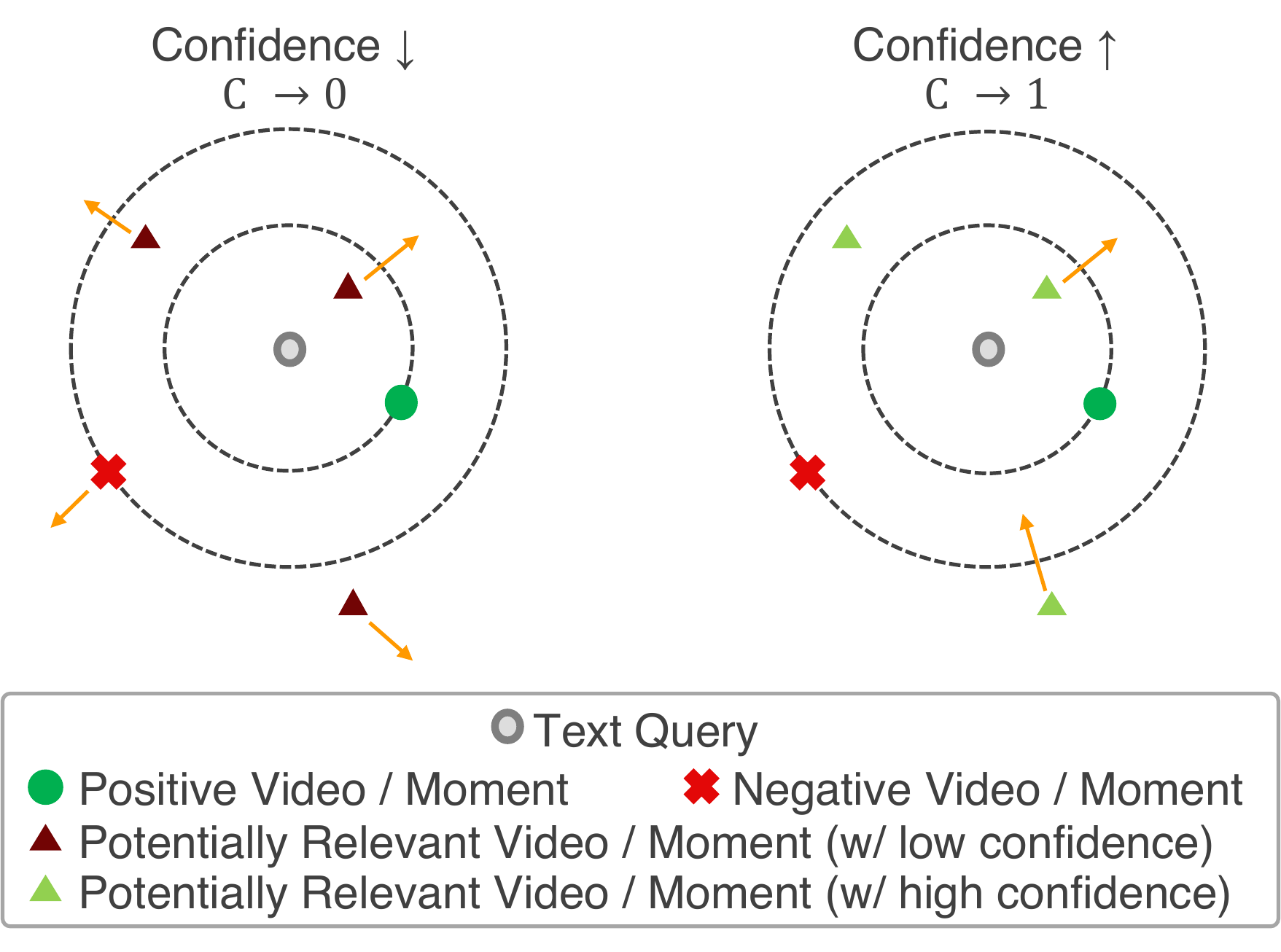}        
  \caption{Illustration of the proposed multi-level ranking loss.
  Here the negative distance is assumed as the relevance score.
  For a potentially relevant sample with a low confidence value, 
  the loss forces the sample further from the text query.
  With regards to a high confidence value, the alternative is enforced as being located within the positive and negative samples.}  
  \label{fig:ranking_loss_overview}
  \end{center}
\end{figure}

\subsection{Loss Function and Potentially Relevant Pair}
In our backbone model XML~\cite{lei2020tvr}, at the time of training,
the loss function is set as a weighted sum of loss functions for $s^{\vr}$ and $s^{\tl}$, respectively,
where the former is a (video-level) ranking loss, similar to \cite{faghri2018vse++}.
The latter is defined as a negative log-likelihood (NLL) loss.
In addition, we add a segment-level ranking loss w.r.t. $s^{\vcmr}$ to optimize the VCMR directly,
but we extend both the ranking loss and NLL loss so that we can handle potentially relevant pairs.

In this paper, we train a model by minimizing a loss function described as 
\begin{equation}    
  \mathcal{L} = 
      \mathcal{L}_{\vr} + \lambda_{\tl}\mathcal{L}_{\tl} + \lambda_{\vcmr}\mathcal{L}_{\vcmr},
  \label{eq:overall_loss}
\end{equation}
where $\mathcal{L}_{\vr}, \mathcal{L}_{\tl},$ and $\mathcal{L}_{\vcmr}$ represent the loss function with respect to $s^{\vr}, s^{\tl},$ and $s^{\vcmr}$,
$\lambda_{\tl}$ and $\lambda_{\vcmr}$ are the weight parameters.

\paragraph{Ranking Loss}
Normal ranking loss considers the relationship between a positive pair and a negative pair.
Using the relevance score $R_{(i,i)}$ with respect to a positive pair $(q_{i}, v_{i})$,
and the relevance scores $R_{(i,\negindex{i})}$ and $R_{(\negindex{j},j)}$ with respect to negative pairs $(q_{i}, v_{\negindex{i}}), (q_{\negindex{j}}, v_{j})$,
following XML~\cite{lei2020tvr}, the normal ranking loss can be formulated as 
\begin{equation}
  \begin{split}
  \mathcal{L}_{\mathrm{pos>neg}} = 
                      \frac{1}{N}[ \sum_{q_{i}}{ [ \Delta + \check{R}_{(i,\negindex{i})} - R_{(i,i)}]_{+}} \\
                      + \sum_{v_{j}}{ [ \Delta + \check{R}_{(\negindex{j},j)} - R_{(j,j)}]_{+}} ],
  \end{split}
  \label{eq:normal_ranking_loss}
\end{equation}
where $N$ is the number of text-video pairs, and $\Delta$ is a margin.
Note that $\check{R}$ randomly samples a negative pair from the top $N_\mathrm{pool} ( 1 \leq N_\mathrm{pool} \leq N )$ items with higher $R$. 
XML defines the video-level ranking loss $\mathcal{L}_{\mathrm{vr}}$ by substituting $s^{\vr}$ into $R$, expecting that the positive pair $s^{\vr}$ becomes larger than that of the negative pair.

In this study, we extend this positive--negative ranking loss to a multilevel ranking loss that includes a potentially relevant pair.
Potentially relevant pairs $(q_{i}, v_{\nonneg{i}}), (q_{\nonneg{j}}, v_{j})$ must hold a higher $R$
than those of the negative pairs $(q_{i}, v_{\negindex{i}}), (q_{\negindex{j}}, v_{j})$,
\ie, $R_{(i,\nonneg{i})} > R_{(i,\negindex{i})}, R_{(\nonneg{j},j)} > R_{(\negindex{j},j)}$.
Moreover, in general, positive pairs $(q_{i}, v_{i})$ and $(q_{j}, v_{j})$ should give a higher $R$ 
than the potentially relevant pairs $(q_{i}, v_{\nonneg{i}})$, $(q_{\nonneg{j}}, v_{j})$, 
yet the potentially relevant pairs could be absolutely positive in terms of semantics.
Therefore, we assume that $R_{(i,i)} \geq R_{(i,\nonneg{i})}, R_{(j,j)} \geq R_{(\nonneg{j},j)}$.
However, at this rate, each potentially relevant pair confidence was not considered.
By incorporating the confidence,
the multi-level ranking loss, that considers the potentially relevant pair together with the confidence, results in the following (also illustrated in \figref{fig:ranking_loss_overview}):
\begin{equation}
  \begin{split}
  \mathcal{L}_{\mathrm{rel > neg}} = \frac{1}{N_\mathrm{pairs}}[ \sum_{v_{\nonneg{i}}}{ [ \Delta + \check{R}_{(i,\negindex{i})} - C_{(i,\nonneg{i})} R_{(i,\nonneg{i})} ]_{+}} \\
                      + \sum_{q_{\nonneg{j}}}{ [ \Delta + \check{R}_{(\negindex{j},j)} - C_{(\nonneg{j},j)} R_{(\nonneg{j},j)} ]_{+}} ].
  \end{split}                        
  \label{eq:nonneg_larger_neg_}
\end{equation}
\begin{equation}
  \begin{split}
  \mathcal{L}_{\mathrm{pos \geq rel}} = \frac{1}{N_\mathrm{pairs}}[ \sum_{v_{\nonneg{i}}}{ [ R_{(i,\nonneg{i})} - C_{(i,\nonneg{i})} R_{(i,i)} ]_{+}} \\
                      + \sum_{q_{\nonneg{j}}}{ [ R_{(\nonneg{j},j)} - C_{(\nonneg{j},j)} R_{(j,j)} ]_{+}} ].
  \end{split}                       
  \label{eq:pos_larger_nonneg_}
\end{equation}
where $N_\mathrm{pairs}$ is the number of the potentially relevant pairs in the batch.
As $C \rightarrow 1$, \equref{eq:nonneg_larger_neg_}, \equref{eq:pos_larger_nonneg_} can be interpreted as a multilevel ranking loss without considering the confidence.
When $C \rightarrow 0$, the potentially relevant pair is almost negative,
and its $R$ is expected to become lower as a result of training.
Let $\lambda_{\mathrm{rel > neg}}^{\vr}$ and $\lambda_{\mathrm{pos \geq rel}}^{\vr}$ be weight parameters,
and by substituting $R=s^{\vr}$, $\mathcal{L}_{\vr}$ is written as 
\begin{equation}    
  \mathcal{L}_{\vr} = 
      \mathcal{L}_{\mathrm{pos > neg}}^{\vr} + \lambda_{\mathrm{rel > neg}}^{\vr}\mathcal{L}_{\mathrm{rel > neg}}^{\vr} + \lambda_{\mathrm{pos \geq rel}}^{\vr}\mathcal{L}_{\mathrm{pos \geq rel}}^{\vr}.
  \label{eq:vr_loss_}
\end{equation}

To define the segment-level ranking loss $\mathcal{L}_{\vcmr}$, we need to choose an appropriate score for substituting $R$.
As per \equref{eq:vcmr_score}, the logarithm of $s^{\vcmr}$ w.r.t. a text query $q_{i}$ and a video segment $v_{j}[t_{\st}^{j}: t_{\ed}^{j}]$,
\ie, $\log{s^{\vcmr}(v_{j},t_{\st}^{j},t_{\ed}^{j}|q_{i},\mathcal{V})} = \log{P_{(i,j),\st}(t_{\st}^{j})} + \log{P_{(i,j),\ed}(t_{\ed}^{j})} + \alpha s^{\vr}(v_{j}|q_{i}))$
can be one of the candidates.
However, $P_{\st}$ and $P_{\ed}$ are normalized using softmax, 
resulting in underestimating nonmaximum values,
making it unsuitable for $R$.
To mitigate this problem, instead of using $P_{\st}(t_{\st}^{j})$ and $P_{\ed}(t_{\ed}^{j})$,
we use their logits (variables before softmax) $o_{\st}(t_{\st}^{j})$ and $o_{\ed}(t_{\ed}^{j})$ to form the VCMR relevance score $R^{\vcmr}$:
\begin{equation}        
  R^{\vcmr}_{(i,j)}= 
      \log{o_{(i,j),\st}(t_{\st}^{j})} + \log{o_{(i,j),\ed}(t_{\ed}^{j})} + \alpha s^{\vr}(v_{j}|q_{i})).
  \label{eq:vcmr_rel_score}
\end{equation}
Note that we add the constraint that $\inf{o_{\st}(t_{\st})} = \inf{o_{\ed}(t_{\ed})} = 10^{-6}$, 
due to the domain of the logarithmic function.
The higher $R^{\vcmr}$ becomes, the higher $s^{\vcmr}$ also becomes.
Using $R^{\vcmr}$, the segment-level ranking loss $\mathcal{L}_{\vcmr}$ can be written as 
\begin{equation}    
  \mathcal{L}_{\vcmr} = 
      \mathcal{L}_{\mathrm{pos > neg}}^{\vcmr} + \lambda_{\mathrm{rel > neg}}^{\vcmr}\mathcal{L}_{\mathrm{rel > neg}}^{\vcmr} + \lambda_{\mathrm{pos \geq rel}}^{\vcmr}\mathcal{L}_{\mathrm{pos \geq rel}}^{\vcmr},
  \label{eq:vcmr_loss_}
\end{equation}
where $\lambda_{\mathrm{rel > neg}}^{\vcmr}$ and $\lambda_\mathrm{{pos \geq rel}}^{\vcmr}$ are weight parameters.

\paragraph{NLL Loss}
\begin{equation}        
  \mathcal{L}_{\mathrm{nll}, 1} =
  -\frac{1}{N}\sum_{(i,i)}[ \log(P_{(i, i),{\st}}(t_{\st}^{i})) + \log(P_{(i,i),{\ed}}(t_{\ed}^{i})) ].
  \label{eq:svmr_loss}
\end{equation}
Based on \equref{eq:svmr_score}, the temporal localization loss is defined as the NLL loss as in \equref{eq:svmr_loss}.
Here, only positive pairs $\{ (q_{i}, v_{i}) \}_{i}$ are considered.
Additionally, we incorporate the potentially relevant pairs into this NLL loss.
The extended loss is as follows:
\begin{equation}        
  \begin{split}  
    \mathcal{L}_{\mathrm{nll},2} = 
   -\frac{1}{N_\mathrm{pairs}}\sum_{(i,\nonneg{i})} C_{(i,\nonneg{i})} [\log(P_{(i,\nonneg{i}),\st}(t_{\st}^{\nonneg{i}})) + \log(P_{(i,\nonneg{i}),\ed}(t_{\ed}^{\nonneg{i}})) ].
  \end{split}
   \label{eq:svmr_loss4non_neg_}
\end{equation}
Because $C \rightarrow 1$, \equref{eq:svmr_loss4non_neg_} works in the same way as in \equref{eq:svmr_loss},
while contributing less as $C \rightarrow 0$,
$\mathcal{L}_{\tl}$ can be formulated as \equref{eq:svmr_loss_total},
where $\lambda_{\tl,2}$ is a weight parameter.
\begin{equation}    
  \mathcal{L}_{\tl} = 
      \mathcal{L}_{\mathrm{nll},1} + \lambda_{\tl,2}\mathcal{L}_{\mathrm{nll},2}.
  \label{eq:svmr_loss_total}
\end{equation}

\section{Experiments}
\subsection{Experimental Setup}
\paragraph{Experiment Overview}
First, we determine the conditions of the potentially relevant pair detector, including the types of $\Phi$ and its threshold $\threshold$.
Second, based on the condition determined above, 
we conducted experiments to determine the effects of the potentially relevant pairs and the use of the confidence score in the loss functions.

For the first stage, we compared three types of conditions: $\Phi_{\mathrm{sent}}$, $\Phi_{\mathrm{vp}}$, and $\Phi_{\mathrm{np-vp}}$.
While $\Phi_{\mathrm{vp}}$ and $\Phi_{\mathrm{np-vp}}$ are described in \secref{sec:detection_and_confidence},
the first one $\Phi_{\mathrm{sent}}$ is a baseline that uses an entire sentence to detect potentially relevant pairs.
This is different from the other two types in the point that these two use parts of the sentences such as VP.
Regarding the threshold $\threshold$ in $\Phi$, we conducted experiments with $\threshold = \{0.5, 0.6, 0.7, 0.8\}$.
These experiments are performed under two types of loss conditions, adding terms that include potentially relevant pairs (PR):
including $\mathcal{L}_{\mathrm{rel > neg}}^{\mathrm{vr}}, \mathcal{L}_{\mathrm{pos \geq rel}}^{\mathrm{vr}}, \mathcal{L}_{\mathrm{nll},2}, \mathcal{L}_{\mathrm{rel > neg}}^{\mathrm{vcmr}}$, and $\mathcal{L}_{\mathrm{pos \geq rel}}^{\mathrm{vcmr}}$, under the following conditions:
(i) the same condition as Lei \etal~\cite{lei2020tvr} \ie, $\mathcal{L}_{\mathrm{pos > neg}}^{\mathrm{vr}} \& \mathcal{L}_{\mathrm{nll},1}$, which we denote as XML,
and (ii) the condition with the segment-level loss (SL) \ie, $\mathcal{L}_{\mathrm{pos > neg}}^{\mathrm{vr}} \& \mathcal{L}_{\mathrm{nll},1} \& \mathcal{L}_{\mathrm{pos > neg}}^{\mathrm{vcmr}}$, which we denote as +SL.
As a result, we obtain two conditions; 
+PR ($\mathcal{L}_{\mathrm{pos>neg}}^{\mathrm{vr}}$ \& $\mathcal{L}_{\mathrm{rel > neg}}^{\mathrm{vr}}$ \& $\mathcal{L}_{\mathrm{pos \geq rel}}^{\mathrm{vr}}$ \& $\mathcal{L}_{\mathrm{nll},1}$ \& $\mathcal{L}_{\mathrm{nll},2}$) 
and +SL+PR ($\mathcal{L}_{\mathrm{pos>neg}}^{\mathrm{vr}}$ \& $\mathcal{L}_{\mathrm{rel > neg}}^{\mathrm{vr}}$ \& $\mathcal{L}_{\mathrm{pos \geq rel}}^{\mathrm{vr}}$ \& $\mathcal{L}_{\mathrm{nll},1}$ \& $\mathcal{L}_{\mathrm{nll},2}$ \& $\mathcal{L}_{\mathrm{pos>neg}}^{\mathrm{vcmr}}$ \& $\mathcal{L}_{\mathrm{rel > neg}}^{\mathrm{vcmr}}$ \& $\mathcal{L}_{\mathrm{pos \geq rel}}^{\mathrm{vcmr}}$).
The weight parameters $\lambda$ are switched on and off \ie, set to one or zero, to achieve the aforementioned conditions,
with the exception of $\lambda_{\mathrm{tl}} = 0.01$, which is essentially constant.

In the second stage, we compare XML, +PR, +SL, and +SL+PR to investigate the effects of +SL and +PR separately.
Moreover, to determine the effect of the confidence score, we conduct experiments by fixing $C\equiv1$ under conditions using the potentially relevant pair.
Consequently, two further conditions are studied: +PR ($C\equiv1$) and +SL+PR ($C\equiv1$).
The condition with $C\equiv1$ can be interpreted as a condition in which all the potentially relevant pairs are treated in the same way.
In the ranking loss, a multi-level ranking loss is applied regardless of each pair relevance,
whereas all the potentially relevant pairs are treated as positive pairs in NLL loss.

\paragraph{Datasets}
To evaluate our method, we conduct experiments on two benchmark datasets for SVMR, 
namely Charades-STA~\cite{gao2017tall} and DiDeMo~\cite{hendricks17iccv}.
These datasets contain untrimmed/unsegmented videos targeting activity videos,
whose segments are paired with text annotations.
Charades-STA is built on top of the Charades~\cite{sigurdsson2016hollywood} dataset.
Videos in Charades were recorded via crowdsourcing; moreover, text annotations were collected semiautomatically from Charades and,
therefore, lack diversity of vocabulary as well as contents compared to the other dataset.
For example, ``{\it person}'' is used to refer to a subject in almost every sentence.
Each video is, on average, 30 s.
The dataset contains 12,408 text-moment pairs for training and 3,720 pairs for the test split.

DiDeMo consists of videos collected from Flickr such that videos with longer than 30 s are trimmed to 30 s;
consequently, most videos hold are 25-30s in length.
Text annotations are collected as referring expressions in each video.
Ground-truth (GT) moment information is given as a five-second-level timestamp,
which is offered by multiple persons and thus,
can vary from person to person.
As per \cite{escorcia2019temporal}, we use a timestamp with at least two persons in agreement as the ground truth.
The dataset was split into 33,005, 4,180, and 4,021 text-moment pairs for training, validation, and testing, respectively.

\paragraph{Evaluation Metrics}
As per prior work \cite{escorcia2019temporal,sudipta2020text,lei2020tvr}, we use recall@$k$ (R@$k$),
indicating the percentage (min:0, max:100) that at least one moment within the top-$k$ items
have a temporal intersection over union (tIoU) with a GT moment larger than a specific threshold.
We report the results with respect to $k=1,10,100$, ${\rm tIoU}=0.5,0.7$ for VCMR.
In addition, for the purpose of analysis, we also report VR R@100 and SVMR R@5 results.
Below, hyphenation after the task name represents tIoU, \eg, VCMR-0.5.

\paragraph{Implementation Details}
The codes were implemented using PyTorch~\cite{NEURIPS2019_9015}, and training was performed using the Adam~\cite{kingma:adam} optimizer.
Specific parameters and settings are based on Lei~\etal~\cite{lei2020tvr}.
Please refer to Sec. A for details.
Note that the XML results were reproduced based on \cite{lei2020tvr} repository\footnote{https://github.com/jayleicn/TVRetrieval}.
As for SBERT, used in $\Phi$ and $\Psi$,
we used the backbone of DistilBERT~\cite{sanh2019distilbert}, which is a memory-light version of BERT,
specifically ``{\it distilbert-base-nli-stsb-mean-tokens}''\footnote{https://github.com/UKPLab/sentence-transformers}.
SBERT embeddings are pre-extracted to compute $\Phi$ and $\Psi$ in a batch at the time of training.

\subsection{Results and Discussions}
\subsubsection{Evaluation of the Potentially Relevant Pair Detector}
\paragraph{Charades-STA}
As per \tabref{tab:charades_vcmr_xml_nonNeg} and \tabref{tab:charades_vcmr_xml_vcmr_nonNeg},
we can observe that, overall, Charades-STA +PR achieves better performance than XML and +SL+PR,
whereas +SL+PR achieved poor performance in terms of the quantitative results compared to XML.
It appears that +SL has a significant influence on the results.
Focusing on the type of the detector, there is no significant difference between $\Phi_{\mathrm{sent}}, \Phi_{\mathrm{vp}}$, and $\Phi_{\mathrm{np-vp}}$ in the Charades-STA.
This may be because of the dataset characteristic---its text annotations are simple in terms of grammatical structures and vocabulary.
However, the best performance is achieved by $\Phi_{\mathrm{np-vp}}$ with $\threshold=0.7$,
showing the effectiveness of $\Phi_{\mathrm{np-vp}}$ to some extent.
We use this condition for the second stage experiment.
The best condition is judged based on the summation of the total scores in \tabref{tab:charades_vcmr_xml_nonNeg} and \tabref{tab:charades_vcmr_xml_vcmr_nonNeg}.

\setlength{\tabcolsep}{1mm}
\begin{table}[t]
   \caption{Comparison of $\Phi_{\mathrm{sent}}, \Phi_{\mathrm{vp}}$, and $\Phi_{\mathrm{np-vp}}$ w.r.t. $\threshold$ under the condition using the potentially relevant pair (+PR) on Charades-STA (test).}
   \centering
   \scalebox{0.8}{   
   \begin{tabular}{cccccccc}\hline
              &              & \multicolumn{3}{c}{VCMR-0.5}                  & \multicolumn{3}{c}{VCMR-0.7}                  \\
              & $\threshold$ & R@1           & R@10          & R@100         & R@1           & R@10          & R@100         \\\hline
   XML~\cite{lei2020tvr}        & ---          & 0.3           & 1.67          & 6.85          & 0.22          & 0.78          & 3.31          \\\hline
   $\Phi_{\mathrm{sent}}$   & 0.5          & \textbf{0.46} & 1.37          & 6.24          & \textbf{0.27} & \textbf{0.81} & 2.85          \\
              & 0.6          & \textbf{0.43} & 1.59          & \textbf{7.15} & \textbf{0.22} & \textbf{0.83} & \textbf{3.44} \\
              & 0.7          & \textbf{0.43} & 1.24          & \textbf{7.55} & \textbf{0.22} & 0.54          & \textbf{3.71} \\
              & 0.8          & \textbf{0.48} & \textbf{1.91} & \textbf{7.58} & \textbf{0.27} & \textbf{0.89} & \textbf{3.95} \\\hline
   $\Phi_{\mathrm{vp}}$         & 0.5          & \textbf{0.51} & \textbf{1.67} & \textbf{6.96} & \textbf{0.3}  & \textbf{0.91} & \textbf{3.44} \\
              & 0.6          & \textbf{0.38} & 1.53          & \textbf{7.02} & \textbf{0.3}  & \textbf{0.81} & \textbf{3.55} \\
              & 0.7          & \textbf{0.43} & \textbf{1.88} & \textbf{7.58} & \textbf{0.27} & \textbf{1.21} & \textbf{3.44} \\
              & 0.8          & \textbf{0.43} & 1.56          & \textbf{7.9}  & \textbf{0.27} & \textbf{0.94} & \textbf{3.95} \\\hline
   $\Phi_{\mathrm{np-vp}}$      & 0.5          & \textbf{0.38} & 1.51          & \textbf{7.31} & \textbf{0.22} & 0.75          & \textbf{3.55} \\
              & 0.6          & \textbf{0.43} & \textbf{1.67} & \textbf{6.91} & \textbf{0.27} & \textbf{0.83} & 3.04          \\
              & 0.7          & \textbf{0.48} & \textbf{1.8}  & \textbf{7.8}  & \textbf{0.32} & \textbf{0.81} & \textbf{4.27} \\
              & 0.8          & \textbf{0.46} & 1.4           & \textbf{7.96} & 0.16          & 0.65          & \textbf{3.66} \\\hline
   \label{tab:charades_vcmr_xml_nonNeg} 
   \end{tabular}
   }
 \end{table}
 
 \setlength{\tabcolsep}{1mm}
 \begin{table}[t]
   \caption{Comparison of $\Phi_{\mathrm{sent}}, \Phi_{\mathrm{vp}}$, and $\Phi_{\mathrm{np-vp}}$ w.r.t. $\threshold$ under the condition using the segment-level loss and the potentially relevant pair (+SL+PR) on Charades-STA (test).}    
   \centering
   \scalebox{0.8}{   
   \begin{tabular}{cccccccc}\hline
                   &              & \multicolumn{3}{c}{VCMR-0.5}                 & \multicolumn{3}{c}{VCMR-0.7}                  \\
                   & $\threshold$ & R@1          & R@10          & R@100         & R@1           & R@10          & R@100         \\\hline
   XML~\cite{lei2020tvr}             & ---          & \textbf{0.3} & \textbf{1.67} & \textbf{6.85} & \textbf{0.22} & \textbf{0.78} & \textbf{3.31} \\\hline
   $\Phi_{\mathrm{sent}}$        & 0.5          & 0.11         & 0.54          & 4.17          & 0.05          & 0.24          & 1.99          \\
                   & 0.6          & 0.11         & 0.65          & 4.3           & 0.03          & 0.3           & 2.15          \\
                   & 0.7          & 0.16         & 0.83          & 4.78          & 0.03          & 0.3           & 2.45          \\
                   & 0.8          & 0.08         & 0.67          & 4.65          & 0.03          & 0.16          & 2.37          \\\hline
   $\Phi_{\mathrm{vp}}$              & 0.5          & 0.13         & 0.43          & 4.49          & 0.11          & 0.16          & 2.1           \\
                   & 0.6          & 0.16         & 0.65          & 4.54          & 0.08          & 0.27          & 2.37          \\
                   & 0.7          & 0.13         & 0.48          & 3.36          & 0.03          & 0.08          & 1.48          \\
                   & 0.8          & 0.18         & 0.89          & 4.78          & 0.08          & 0.46          & 2.2           \\\hline
   $\Phi_{\mathrm{np-vp}}$           & 0.5          & 0.13         & 0.43          & 2.39          & 0.08          & 0.13          & 1.13          \\
                   & 0.6          & 0.08         & 0.19          & 0.62          & 0.08          & 0.13          & 0.27          \\
                   & 0.7          & 0.11         & 0.59          & 4.68          & 0.03          & 0.22          & 2.26          \\
                   & 0.8          & 0.11         & 0.27          & 1.51          & 0.0           & 0.05          & 0.73          \\\hline
   \label{tab:charades_vcmr_xml_vcmr_nonNeg}                   
   \end{tabular}
   }
 \end{table}

 \paragraph{DiDeMo}
 As per \tabref{tab:didemo_vcmr_xml_nonNeg} and \tabref{tab:didemo_vcmr_xml_vcmr_nonNeg},
 unlike Charades-STA, we can see that +SL+PR is better than +PR overall,
 suggesting that +SL has a positive effect on the performance in DiDeMo.
 \tabref{tab:didemo_vcmr_xml_vcmr_nonNeg} indicates that $\Phi_{\mathrm{sent}}$, which uses the entire sentence, is insufficient to detect the potentially relevant pair, 
 whereas both $\Phi_{\mathrm{vp}}$ and $\Phi_{\mathrm{np-vp}}$ basically achieved higher scores than XML.
 Thus, proper detection of the potentially relevant pair seems necessary.
 As with Charades-STA, we adopt +SL+PR with $\Phi_{\mathrm{vp}}, \threshold=0.5$ for the second-stage experiment.
 
 \setlength{\tabcolsep}{1mm}
 \begin{table}[t]
  \caption{Comparison of $\Phi_{\mathrm{sent}}, \Phi_{\mathrm{vp}}$, and $\Phi_{\mathrm{np-vp}}$ w.r.t. $\threshold$ under the condition using the potentially relevant pair (+PR) on DiDeMo (val).}    
  \centering
  \scalebox{0.8}{   
  \begin{tabular}{cccccccc}\hline
             &              & \multicolumn{3}{c}{VCMR-0.5} & \multicolumn{3}{c}{VCMR-0.7} \\
             & $\threshold$ & R@1           & R@10 & R@100 & R@1           & R@10 & R@100 \\\hline
  XML~\cite{lei2020tvr}        & ---          & 1.87          & 8.95 & 32.51 & 0.57          & 3.52 & 18.49 \\\hline
  $\Phi_{\mathrm{sent}}$   & 0.5          & 1.6           & 6.29 & 29.67 & 0.55          & 2.56 & 15.26 \\
             & 0.6          & 1.39          & 5.02 & 25.24 & \textbf{0.72} & 2.37 & 13.13 \\
             & 0.7          & 1.67          & 6.58 & 27.92 & \textbf{0.79} & 2.61 & 14.62 \\
             & 0.8          & 1.7           & 6.39 & 28.83 & \textbf{0.72} & 2.39 & 14.33 \\\hline
  $\Phi_{\mathrm{vp}}$         & 0.5          & \textbf{1.94} & 7.11 & 32.03 & \textbf{0.89} & 3.16 & 16.12 \\
             & 0.6          & \textbf{1.94} & 6.41 & 28.64 & \textbf{0.74} & 2.68 & 16.0  \\
             & 0.7          & \textbf{1.91} & 7.13 & 32.08 & \textbf{0.57} & 2.82 & 16.58 \\
             & 0.8          & 1.7           & 5.62 & 27.85 & \textbf{0.79} & 2.49 & 13.9  \\\hline
  $\Phi_{\mathrm{np-vp}}$      & 0.5          & 1.77          & 6.51 & 28.54 & \textbf{0.62} & 3.11 & 15.31 \\
             & 0.6          & 1.63          & 6.03 & 29.23 & \textbf{0.77} & 2.51 & 15.72 \\
             & 0.7          & 1.77          & 7.3  & 29.64 & 0.26          & 2.22 & 14.21 \\
             & 0.8          & 1.84          & 6.7  & 30.5  & \textbf{0.62} & 2.68 & 16.08 \\\hline
  \end{tabular}
  }
  \label{tab:didemo_vcmr_xml_nonNeg} 
\end{table}

\setlength{\tabcolsep}{1mm}
\begin{table}[t]
  \caption{Comparison of $\Phi_{\mathrm{sent}}, \Phi_{\mathrm{vp}}$, and $\Phi_{\mathrm{np-vp}}$ w.r.t. $\threshold$ under the condition using the segment-level loss and the potentially relevant pair (+SL+PR) on DiDeMo (val).}    
  \centering
  \scalebox{0.8}{   
  \begin{tabular}{cccccccc}\hline
             &              & \multicolumn{3}{c}{VCMR-0.5}                    & \multicolumn{3}{c}{VCMR-0.7}                   \\
             & $\threshold$ & R@1           & R@10           & R@100          & R@1           & R@10          & R@100          \\\hline
  XML~\cite{lei2020tvr}        & ---          & 1.87          & 8.95           & 32.51          & 0.57          & 3.52          & 18.49          \\\hline
  $\Phi_{\mathrm{sent}}$   & 0.5          & 1.67          & 7.3            & 28.47          & 0.55          & 2.94          & 16.39          \\
             & 0.6          & \textbf{1.89} & 7.49           & 30.41          & 0.77          & 2.61          & 16.41          \\
             & 0.7          & 1.6           & 7.08           & 28.06          & 0.62          & 2.34          & 13.28          \\
             & 0.8          & 1.6           & 7.49           & 29.55          & 0.57          & 2.54          & 13.61          \\\hline
  $\Phi_{\mathrm{vp}}$         & 0.5          & \textbf{2.89} & \textbf{10.93} & \textbf{37.97} & \textbf{0.77} & \textbf{4.93} & \textbf{24.23} \\
             & 0.6          & \textbf{2.46} & \textbf{9.52}  & \textbf{34.57} & \textbf{0.65} & \textbf{3.71} & \textbf{20.74} \\
             & 0.7          & \textbf{2.08} & \textbf{10.07} & \textbf{35.81} & \textbf{0.65} & 3.47          & \textbf{19.78} \\
             & 0.8          & \textbf{2.51} & \textbf{10.65} & \textbf{37.46} & 0.5           & \textbf{3.83} & \textbf{21.89} \\\hline
  $\Phi_{\mathrm{np-vp}}$      & 0.5          & \textbf{2.7}  & \textbf{10.33} & \textbf{36.27} & \textbf{0.72} & \textbf{5.12} & \textbf{24.78} \\
             & 0.6          & \textbf{2.94} & \textbf{10.79} & \textbf{37.18} & \textbf{0.67} & \textbf{4.67} & \textbf{22.46} \\
             & 0.7          & \textbf{2.22} & \textbf{9.43}  & \textbf{36.2}  & 0.43          & 3.33          & \textbf{19.5}  \\
             & 0.8          & \textbf{1.96} & \textbf{10.17} & \textbf{35.1}  & 0.45          & 3.21          & 18.09          \\\hline
  \label{tab:didemo_vcmr_xml_vcmr_nonNeg} 
  \end{tabular}
  }
\end{table}

 \subsubsection{Evaluation of the Proposed Loss}
 \paragraph{The annalysis of VCMR performance}
  Overall, \tabref{tab:charades_vcmr_ablation} and \tabref{tab:didemo_vcmr_ablation} show that $C\equiv1$ damages the performance under all conditions involved with the potentially relevant pair.
  This indicates that the appropriate design of each potentially relevant pair confidence is important.
  As for Charades-STA, the significant drop in score can be observed under +SL.
  We can confirm that the lower scores of +SL+PR in \tabref{tab:charades_vcmr_xml_vcmr_nonNeg} are triggered by segment-level loss.  
  On the other hand, DiDeMo best scores are achieved under +SL+PR.
  Because +SL and +PR do not reach the scores of +SL+PR, proper design of loss types seems crucial.  
  In both datasets we can confirm, with the appropriate loss functions that incorporate each potentially relevant pair effectively,
  that retrieval performance can be enhanced.  

 \paragraph{The effect of VR and SVMR on VCMR performance}
 Under $C\equiv1$, we can observe that SVMR and VR basically deteriorates,
 which appears to lead to the drop in the VCMR performance.
 Looking at the conditions with higher scores, 
 we can see that the increase in the SVMR (temporal localization) scores, mainly contributed to the VCMR performance improvement.
 Training with potentially relevant pairs might have a data-augmentation-like effect and,
 as a result, succeeded in text-moment grounding for difficult samples.

  \setlength{\tabcolsep}{1.0pt}
  \begin{table}[t]
   \caption{Evaluation of the proposed loss on Charades-STA (test) under $\Phi_{\mathrm{np-vp}}, \threshold=0.7$.}  
   \centering  
   \scalebox{0.78}{   
   \begin{tabular}{ccccccccccc}\hline
                                                    &              & \multicolumn{3}{c}{VCMR-0.5}                & \multicolumn{3}{c}{VCMR-0.7}                  & SVMR-0.5       & SVMR-0.7        & VR   \\
                                                    & $\threshold$ & R@1           & R@10         & R@100        & R@1           & R@10          & R@100         & R@5            & R@5             & R@100\\\hline
   XML~\cite{lei2020tvr}                            & ---          & 0.3           & 1.67         & 6.85         & 0.22          & 0.78          & 3.31          & 51.75          & 27.85           & 26.59\\\hline
   +PR                                              & 0.7          & \textbf{0.48} & \textbf{1.8} & \textbf{7.8} & \textbf{0.32} & \textbf{0.81} & \textbf{4.27} & \textbf{55.99} & \textbf{29.62}  & 25.89\\\hline
   +PR ($C\equiv1$)                                 & 0.7          & \textbf{0.38} & 1.24         & 6.48         & \textbf{0.22} & 0.59          & 2.96          & \textbf{51.83} & 27.47           & 22.58\\\hline
   +SL                                              & ---          & 0.11          & 0.35         & 3.25         & 0.0           & 0.11          & 1.16          & 48.6           & 25.46           & 23.74\\\hline  
   +SL+PR                                           & 0.7          & 0.11          & 0.59         & 4.68         & 0.03          & 0.22          & 2.26          & 51.24          & 27.63           & 21.96\\\hline
   \multicolumn{1}{l}{+SL+PR ($C\equiv1$)}          & 0.7          & 0.11          & 0.3          & 2.74         & 0.0           & 0.03          & 0.7           & 38.95          & 17.02           & 19.73\\\hline
   \end{tabular}  
   }
   \label{tab:charades_vcmr_ablation}                   
 \end{table}

 \setlength{\tabcolsep}{1.0pt}
 \begin{table}[t]
   \centering
   \caption{Evaluation of the proposed loss on DiDeMo (test) under $\Phi_{\mathrm{vp}}, \threshold=0.5$.}     
   \scalebox{0.78}{   
   \begin{tabular}{ccccccccccc}\hline
                                &              & \multicolumn{3}{c}{VCMR-0.5}                    & \multicolumn{3}{c}{VCMR-0.7}                   & SVMR-0.5       & SVMR-0.7        & VR            \\                                
                                & $\threshold$ & R@1           & R@10           & R@100          & R@1           & R@10          & R@100          & R@5            & R@5             & R@100         \\\hline   
   XML~\cite{lei2020tvr}        & ---          & 1.74          & 9.15           & 32.26          & 0.65          & 3.61          & 17.36          & 82.07          & 56.83           & 71.23         \\\hline
   +PR                          & 0.5          & 1.47          & 6.89           & 31.01          & 0.57          & 2.66          & 15.27          & 81.97          & 52.1            & \textbf{73.19}\\\hline
   +PR ($C\equiv1$)             & 0.5          & 1.09          & 6.12           & 24.97          & 0.4           & 2.76          & 15.34          & 77.07          & 53.49           & 65.41         \\\hline
   +SL                          & ---          & 1.74          & 8.6            & \textbf{33.42} & 0.6           & 3.36          & \textbf{18.43} & 80.85          & 51.85           & 70.55         \\\hline  
   +SL+PR                       & 0.5          & \textbf{2.21} & \textbf{11.34} & \textbf{37.88} & \textbf{0.6}  & \textbf{5.05} & \textbf{23.78} & \textbf{86.69} & 53.17           & \textbf{71.33}\\\hline
   +SL+PR ($C\equiv1$)          & 0.5          & 1.49          & 4.92           & 20.72          & 0.42          & 3.28          & 15.62          & 76.42          & 51.58           & 60.38         \\\hline
 
   \end{tabular}
   }
   \label{tab:didemo_vcmr_ablation}                   
 \end{table}

\subsection{Qualitative Results}
\label{sec:qualitative_results}
\begin{figure*}
  \begin{center}    
   \includegraphics[width=0.78\linewidth]{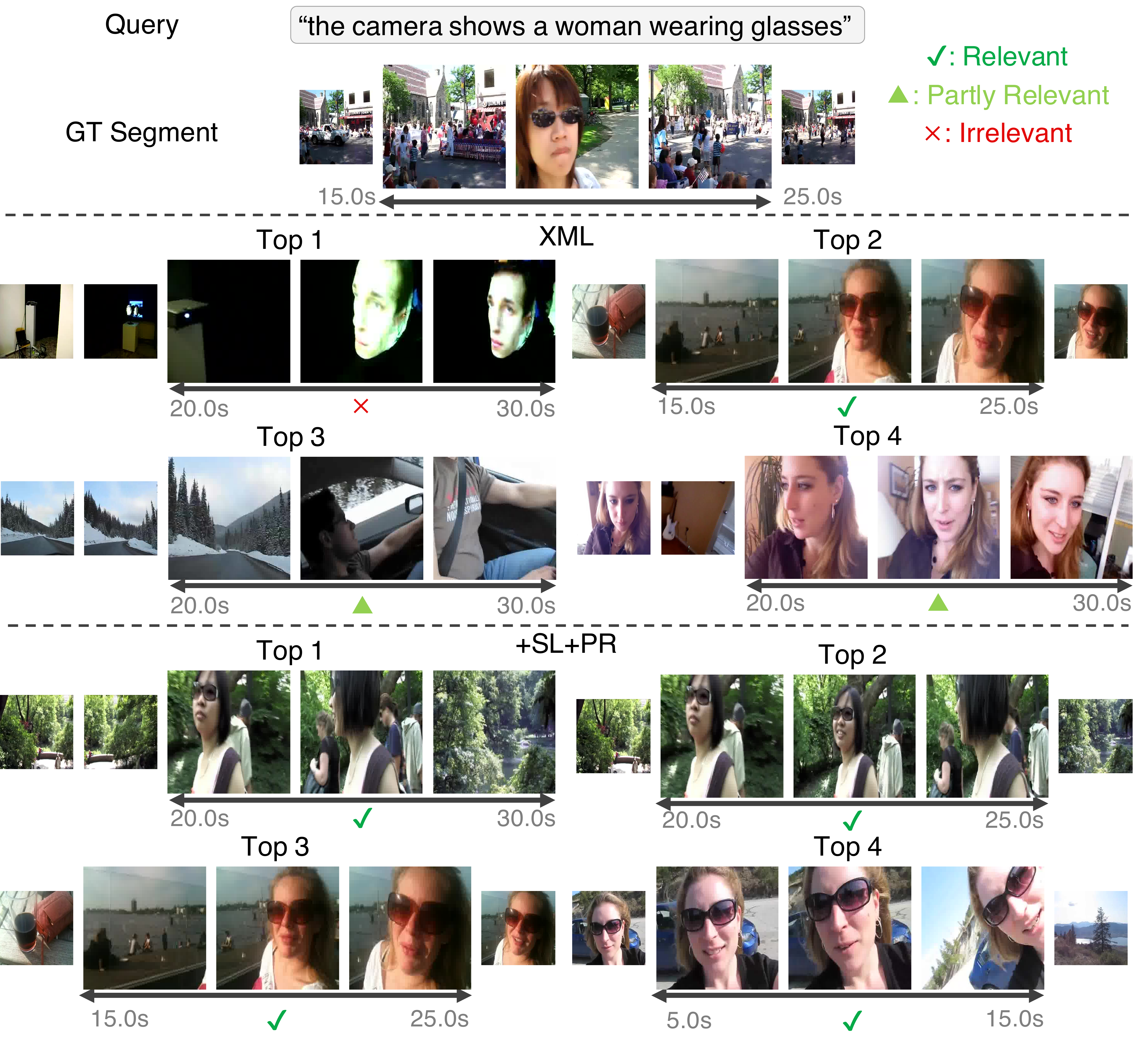}        
  \caption{Example of the qualitative results on DiDeMo (test) under $\Phi_\mathrm{vp}$, $\threshold=0.5$.}  
  \label{fig:didemo_qual_result3}
  \end{center}
\end{figure*}

\figref{fig:didemo_qual_result3} is an example of the qualitative results on DiDeMo showing up to top-4 candidate moments.
Although XML certainly outputs face-enhanced moments, which are somewhat relevant to the query,
the output moments vary in terms of semantics.
For example, while the ``\textit{man}'' in Top 3 is ``\textit{wearing glasses},''
the ``\textit{woman}'' in Top 4 is not ``\textit{wearing glasses}.''
Moreover, Top 1 does not include ``\textit{woman}'' or ``\textit{glasses}''.

On the other hand, +SL+PR provides relevant moments to the query even though they are not the ground-truth moments.
This tendency can be explained by the confidence design with SBERT 
as SBERT can compute semantic-sensitive similarity between sentences.
Thus, we can confirm that the effective use of a potentially relevant pair can lead to more relevant moment outputs.
Please refer to Sec. C for more qualitative results.

\section{Conclusion}
In this work, which is focused on training a model for video corpus moment retrieval,
we proposed a novel training method that takes advantage of potentially relevant pairs,
allowing for consideration of the many-to-many correspondences between text and video moments.
We performed linguistic analysis using SentenceBERT to detect the potentially relevant pairs,
then incorporated them into loss functions effectively.
As a result, the retrieval performance was more relevant to text queries and achieved better quantitative results.
Future work includes designing a more sophisticated potentially relevant pair detector and confidence computation function.

\begin{acks}  
  This work was supported by JSPS KAKENHI Grant Number JP19H01115, JP20H05556, and Moonshot R\&D Grant Number JPMJPS2011.
  The authors would like to thank Yuki Kawana, Kohei Uehara, and Hiroaki Yamane for helpful discussions, and Yusuke Mori for technical help.
\end{acks}

\appendix

\section{Implementation Details}
\label{sec:supp_implementation_details}
The text encoder RoBERTa~\cite{liu2019roberta} was finetuned using each dataset beforehand, with the masked language modeling~\cite{devlin-etal-2019-bert}.
For the video representation, as per Lei~\etal~\cite{lei2020tvr}, 
we used the concatenation of clip-level ResNet-152~\cite{he2016deep} features and temporal endpoint features~\cite{hendricks17iccv}.
Note that frames are obtained at a rate of 3 fps.
As for the feature dimensions, the word embedding size $D_{q}=768$,
the total size of the video representation $D_{v}$ is 2050,
and the common hidden feature $d=256$.
The maximum word length $l_{q}=30$, 
and the maximum clip length $l_{v}$ is 20 for Charades-STA, 12 for DiDeMo.
Note that the clip length $\delta t$ in each dataset is $\delta t = 3.0$ second in Charades-STA and $\delta t = 2.5$ second in DiDeMo, respectively.
The constraint parameters in \equref{eq:inference_constraints} are $n_{\mathrm{min}}=1, n_{\mathrm{max}}=8$ in Charades-STA, $n_{\mathrm{min}}=2, n_{\mathrm{max}}=4$ in DiDeMo.

For optimization, we used Adam~\cite{kingma:adam} with an initial learning rate of 0.0001.
The batch size $N$ is 128.
We trained 100 epochs for XML~\cite{lei2020tvr}, while 200 epochs for other conditions.
In the first 20 epochs, $N_{\mathrm{pool}}$ is set to 1, afterwards $N_{\mathrm{pool}}=20$.
The margin $\Delta=0.1$, the weight parameter $\alpha=20$.

\section{Additional Experiments}
In addition to the experiments in the main paper, 
to confirm if it is reasonable to incorporate all the loss terms involved with the potentially relevant pair at the same time,
we conducted ablation study about loss terms by excluding the following loss terms,
\ie, $\mathcal{L}_{\mathrm{rel > neg}}^{\mathrm{vr}}, \mathcal{L}_{\mathrm{pos \geq rel}}^{\mathrm{vr}}, \mathcal{L}_{\mathrm{nll},2}, \mathcal{L}_{\mathrm{rel > neg}}^{\mathrm{vcmr}}$, and $\mathcal{L}_{\mathrm{pos \geq rel}}^{\mathrm{vcmr}}$.
Based on the conditions in the main paper 
+PR ($\mathcal{L}_{\mathrm{pos>neg}}^{\mathrm{vr}}$ \& $\mathcal{L}_{\mathrm{rel > neg}}^{\mathrm{vr}}$ \& $\mathcal{L}_{\mathrm{pos \geq rel}}^{\mathrm{vr}}$ \& $\mathcal{L}_{\mathrm{nll},1}$ \& $\mathcal{L}_{\mathrm{nll},2}$)
and +SL+PR ($\mathcal{L}_{\mathrm{pos>neg}}^{\mathrm{vr}}$ \& $\mathcal{L}_{\mathrm{rel > neg}}^{\mathrm{vr}}$ \& $\mathcal{L}_{\mathrm{pos \geq rel}}^{\mathrm{vr}}$ \& $\mathcal{L}_{\mathrm{nll},1}$ \& $\mathcal{L}_{\mathrm{nll},2}$ \& $\mathcal{L}_{\mathrm{pos>neg}}^{\mathrm{vcmr}}$ \& $\mathcal{L}_{\mathrm{rel > neg}}^{\mathrm{vcmr}}$ \& $\mathcal{L}_{\mathrm{pos \geq rel}}^{\mathrm{vcmr}}$),
we excluded loss terms, which are involved with the potentially relevant pair,
depending on the loss types.
For example, for the ranking loss, 
$\mathcal{L}_{\mathrm{rel > neg}}^{\mathrm{vr}}$ and $\mathcal{L}_{\mathrm{pos \geq rel}}^{\mathrm{vr}}$ ($\mathcal{L}_{\mathrm{rel > neg}}^{\mathrm{vcmr}}$ and $\mathcal{L}_{\mathrm{pos \geq rel}}^{\mathrm{vcmr}}$)
are excluded at the same time,
while $\mathcal{L}_{\mathrm{nll}, 2}$ is excluded for the NLL loss.
In each condition, we also conducted experiments by fixing $C\equiv1$ to see the effect of the confidence design as in the main paper.

\tabref{tab:charades-sta_ablation_study_loss_terms} and \tabref{tab:didemo_ablation_study_loss_terms} are the experiment results.
As pointed out in the main paper, we can see that $C\equiv1$ basically degrades the performance,
showing the importance of the confidence design.
In addition, as mentioned in the paper, 
we can observe that the SVMR-0.5 improvement mainly contributes to the VCMR better results.
Moreover, based on the summation of total VCMR scores as in the main paper,
we can say that including all the loss terms (+PR or +SL+PR) is suitable, because this condition remarks the best or comparable performance.

\setlength{\tabcolsep}{1mm}
\begin{table*}[t]  
  \caption{Ablation study with respect to loss terms and the evaluation of the confidence design on Charades-STA (test) under $\Phi_{\mathrm{np-vp}}$, $\theta=0.7$.
  \checkmark represents that we adopt the corresponding condition.}
  \centering
  \scalebox{0.85}{ 
    \label{tab:charades-sta_ablation_study_loss_terms}    
  \begin{tabular}{c|ccccc|c|ccccccccc} \hline
                          &                               &                               &                                &                                 &                                 &            & \multicolumn{3}{c}{VCMR-0.5}                            & \multicolumn{3}{c}{VCMR-0.7}                            & SVMR-0.5          & SVMR-0.7          & VR             \\ 
                          & $\mathcal{L}_{\mathrm{rel > neg}}^{\vr}$ & $\mathcal{L}_{\mathrm{pos \geq rel}}^{\vr}$ & $\mathcal{L}_{\mathrm{nll},2}$ & $\mathcal{L}_{\mathrm{rel > neg}}^{\vcmr}$ & $\mathcal{L}_{\mathrm{pos \geq rel}}^{\vcmr}$ & $C\equiv1$ & R@1              & R@10             & R@100             & R@1              & R@10             & R@100             & R@5               & R@5               & R@100          \\ \hline                          
  XML~\cite{lei2020tvr}   & ---                           & ---                           & ---                            & ---                             & ---                             & ---        & 0.3              & 1.67             & 6.85              & 0.22             & 0.78             & 3.31              & 51.75             & 27.85             & 26.59          \\ \hline  
  \multirow{6}{*}{$\mathcal{L}_{\mathrm{pos > neg}}^{\vr}$ \& $\mathcal{L}_{\mathrm{nll},1}$}    
                          & \checkmark                    & \checkmark                    & \checkmark                     & ---                             & ---                             &            & \textbf{0.48}    & \textbf{1.8}     & \textbf{7.8}      & \textbf{0.32}    & \textbf{0.81}    & \textbf{4.27}     & \textbf{55.99}    & \textbf{29.62}    & 25.89          \\ 
                          & \checkmark                    & \checkmark                    & \checkmark                     & ---                             & ---                             & \checkmark & \textbf{0.38}    & 1.24             & 6.48              & \textbf{0.22}    & 0.59             & 2.96              & \textbf{51.83}    & 27.47             & 22.58          \\
                          &                               &                               & \checkmark                     & ---                             & ---                             &            & \textbf{0.56}    & 1.51             & 6.56              & \textbf{0.27}    & \textbf{0.81}    & 3.17              & \textbf{53.01}    & 26.8              & \textbf{28.71} \\
                          &                               &                               & \checkmark                     & ---                             & ---                             & \checkmark & \textbf{0.65}    & \textbf{1.94}    & \textbf{7.12}     & \textbf{0.43}    & \textbf{0.91}    & \textbf{3.92}     & \textbf{54.87}    & 26.72             & \textbf{28.52} \\
                          & \checkmark                    & \checkmark                    &                                & ---                             & ---                             &            & \textbf{0.46}    & 1.42             & \textbf{6.85}     & \textbf{0.32}    & 0.73             & \textbf{3.33}     & \textbf{54.78}    & \textbf{29.14}    & 25.22          \\
                          & \checkmark                    & \checkmark                    &                                & ---                             & ---                             & \checkmark & 0.4              & 1.18             & 5.78              & \textbf{0.27}    & 0.64             & 3.04              & \textbf{52.8}     & \textbf{28.25}    & 22.63          \\ \hline
  \multirow{8}{*}{$\mathcal{L}_{\mathrm{pos > neg}}^{\vr}$ \& $\mathcal{L}_{\mathrm{nll},1}$ \& $\mathcal{L}_{\mathrm{pos > neg}}^{\vcmr}$}
                          & \checkmark                    & \checkmark                    & \checkmark                     & \checkmark                      & \checkmark                      &            & 0.11             & 0.59             & 4.68              & 0.03             & 0.22             & 2.26              & 51.24             & 27.63             & 21.96          \\ 
                          & \checkmark                    & \checkmark                    & \checkmark                     & \checkmark                      & \checkmark                      & \checkmark & 0.11             & 0.3              & 2.74              & 0.0              & 0.03             & 0.7               & 38.95             & 17.02             & 19.73          \\
                          &                               &                               & \checkmark                     & \checkmark                      & \checkmark                      &            & 0.16             & 0.78             & 4.87              & 0.11             & 0.4              & 2.28              & 48.09             & 24.89             & 22.63          \\
                          &                               &                               & \checkmark                     & \checkmark                      & \checkmark                      & \checkmark & 0.11             & 0.46             & 2.37              & 0.08             & 0.27             & 0.99              & 43.04             & 20.89             & 16.45          \\
                          & \checkmark                    & \checkmark                    &                                & \checkmark                      & \checkmark                      &            & 0.08             & 0.56             & 4.27              & 0.03             & 0.11             & 2.15              & 51.05             & 27.39             & 23.17          \\
                          & \checkmark                    & \checkmark                    &                                & \checkmark                      & \checkmark                      & \checkmark & 0.13             & 0.67             & 3.2               & 0.05             & 0.27             & 1.56              & 44.46             & 22.34             & 17.34          \\
                          & \checkmark                    & \checkmark                    & \checkmark                     &                                 &                                 &            & 0.16             & 1.02             & 4.27              & 0.11             & 0.48             & 2.31              & 51.1              & 27.28             & 22.02          \\
                          & \checkmark                    & \checkmark                    & \checkmark                     &                                 &                                 & \checkmark & 0.22             & 0.7              & 4.68              & 0.03             & 0.13             & 1.91              & 47.1              & 21.83             & 24.38          \\ \hline
  \end{tabular}
  }
\end{table*}

\setlength{\tabcolsep}{1mm}
\begin{table*}[t]  
  \caption{Ablation study with respect to loss terms and the evaluation of the confidence design on DiDeMo (test) under $\Phi_{\mathrm{vp}}$, $\theta=0.5$
  \checkmark represents that we adopt the corresponding condition.}
  \centering
  \scalebox{0.85}{  
  \label{tab:didemo_ablation_study_loss_terms}    
  \begin{tabular}{c|ccccc|c|ccccccccc} \hline
                                   &                               &                               &                                &                                 &                                 &            & \multicolumn{3}{c}{VCMR-0.5}                            & \multicolumn{3}{c}{VCMR-0.7}                            & SVMR-0.5          & SVMR-0.7          & VR             \\                                   
                                   & $\mathcal{L}_{\mathrm{rel > neg}}^{\vr}$ & $\mathcal{L}_{\mathrm{pos \geq rel}}^{\vr}$ & $\mathcal{L}_{\mathrm{nll},2}$ & $\mathcal{L}_{\mathrm{rel > neg}}^{\vcmr}$ & $\mathcal{L}_{\mathrm{pos \geq rel}}^{\vcmr}$ & $C\equiv1$ & R@1              & R@10             & R@100             & R@1              & R@10             & R@100             & R@5               & R@5               & R@100          \\ \hline
  XML~\cite{lei2020tvr}            & ---                           & ---                           & ---                            & ---                             & ---                             & ---        & 1.74             & 9.15             & 32.26             & 0.65             & 3.61             & 17.36             & 82.07             & 56.83             & 71.23          \\ \hline
  \multirow{6}{*}{$\mathcal{L}_{\mathrm{pos > neg}}^{\vr}$ \& $\mathcal{L}_{\mathrm{nll},1}$}        
                                   & \checkmark                    & \checkmark                    & \checkmark                     & ---                             & ---                             &            & 1.47             & 6.89             & 31.01             & 0.57             & 2.66             & 15.27             & 81.97             & 52.1              & \textbf{73.19} \\ 
                                   & \checkmark                    & \checkmark                    & \checkmark                     & ---                             & ---                             & \checkmark & 1.09             & 6.12             & 24.97             & 0.4              & 2.76             & 15.34             & 77.07             & 53.49             & 65.41          \\
                                   &                               &                               & \checkmark                     & ---                             & ---                             &            & \textbf{1.99}    & 8.03             & \textbf{33.77}    & \textbf{0.75}    & 3.31             & \textbf{20.54}    & 81.4              & \textbf{56.9}     & \textbf{72.77} \\
                                   &                               &                               & \checkmark                     & ---                             & ---                             & \checkmark & 1.52             & 7.96             & 29.74             & 0.57             & \textbf{3.78}    & \textbf{18.78}    & 78.61             & 53.69             & \textbf{71.23} \\
                                   & \checkmark                    & \checkmark                    &                                & ---                             & ---                             &            & 1.19             & 6.67             & 30.37             & 0.57             & 2.93             & 14.75             & 80.73             & 53.02             & \textbf{72.99} \\
                                   & \checkmark                    & \checkmark                    &                                & ---                             & ---                             & \checkmark & 1.52             & 7.34             & 27.43             & 0.37             & 2.69             & 13.93             & 79.76             & 51.16             & 66.58          \\ \hline
                                   \multirow{8}{*}{$\mathcal{L}_{\mathrm{pos > neg}}^{\vr}$ \& $\mathcal{L}_{\mathrm{nll},1}$ \& $\mathcal{L}_{\mathrm{pos > neg}}^{\vcmr}$}                                  
                                   & \checkmark                    & \checkmark                    & \checkmark                     & \checkmark                      & \checkmark                      &            & \textbf{2.21}    & \textbf{11.34}   & \textbf{37.88}    & 0.6              & \textbf{5.05}    & \textbf{23.78}    & \textbf{86.69}    & 53.17             & \textbf{71.33} \\    
                                   & \checkmark                    & \checkmark                    & \checkmark                     & \checkmark                      & \checkmark                      & \checkmark & 1.49             & 4.92             & 20.72             & 0.42             & 3.28             & 15.62             & 76.42             & 51.58             & 60.38          \\
                                   &                               &                               & \checkmark                     & \checkmark                      & \checkmark                      &            & \textbf{2.66}    & 8.46             & \textbf{33.15}    & \textbf{0.8}     & \textbf{6.32}    & \textbf{26.83}    & \textbf{83.96}    & \textbf{58.97}    & \textbf{72.12} \\
                                   &                               &                               & \checkmark                     & \checkmark                      & \checkmark                      & \checkmark & 1.27             & 5.17             & 22.03             & 0.25             & 3.26             & 16.34             & 75.95             & 51.13             & 62.52          \\
                                   & \checkmark                    & \checkmark                    &                                & \checkmark                      & \checkmark                      &            & \textbf{2.41}    & \textbf{9.35}    & \textbf{34.67}    & \textbf{0.75}    & \textbf{6.19}    & \textbf{26.16}    & \textbf{82.34}    & \textbf{58.79}    & \textbf{71.47} \\
                                   & \checkmark                    & \checkmark                    &                                & \checkmark                      & \checkmark                      & \checkmark & 0.62             & 3.48             & 18.01             & 0.15             & 1.99             & 13.5              & 74.76             & 48.77             & 58.22          \\
                                   & \checkmark                    & \checkmark                    & \checkmark                     &                                 &                                 &            & 1.39             & 7.61             & 30.91             & 0.5              & 3.08             & 15.87             & 81.37             & 52.97             & 71.1           \\
                                   & \checkmark                    & \checkmark                    & \checkmark                     &                                 &                                 & \checkmark & 1.29             & 7.81             & 31.31             & 0.37             & 2.76             & 14.08             & \textbf{84.51}    & 45.26             & 67.99          \\ \hline
  \end{tabular}
  }
\end{table*}

\section{Qualitative Results}
\label{sec:supp_qualitative_results}
\begin{figure*}
  \begin{center}    
   \includegraphics[width=0.9\linewidth]{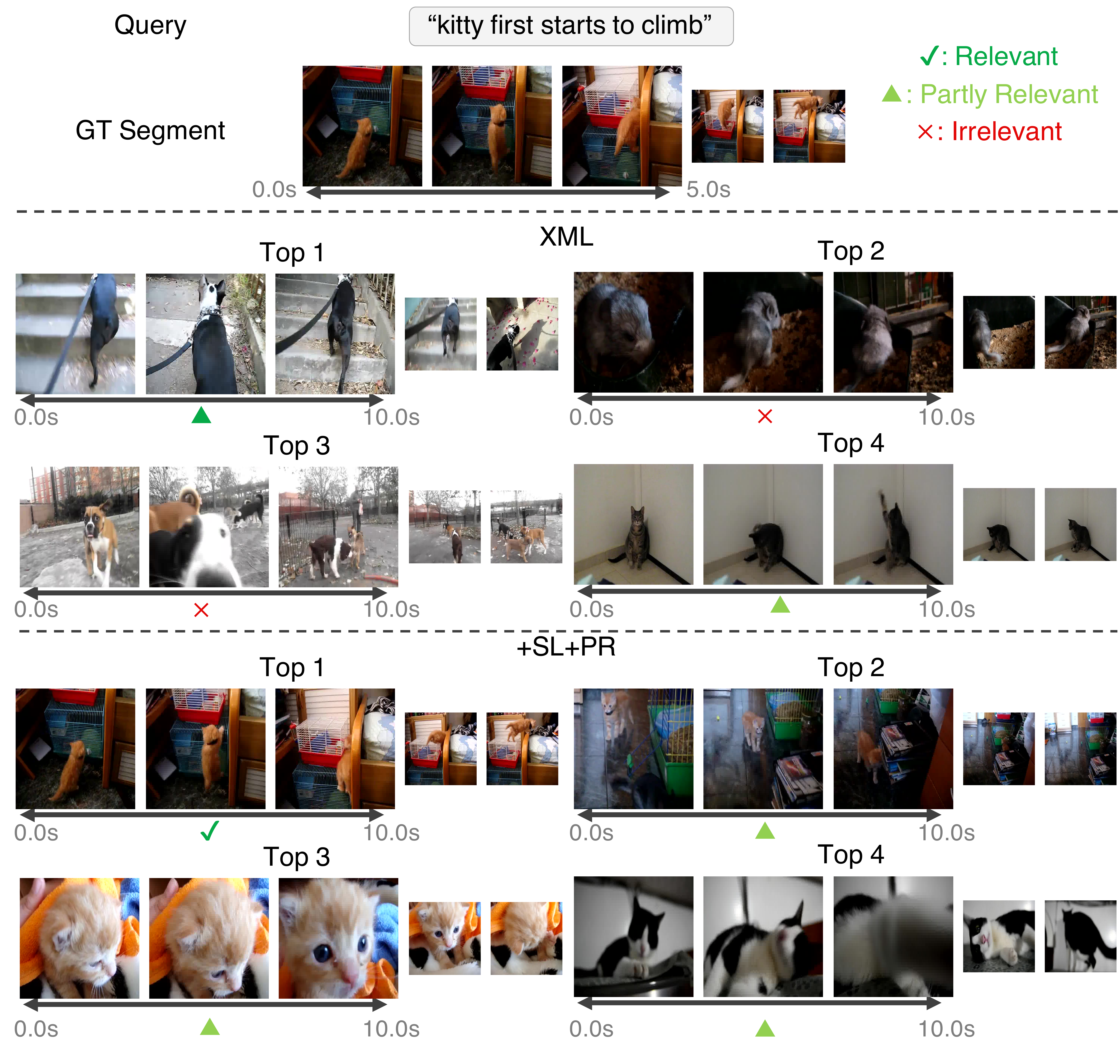}        
  \caption{Example of the qualitative results on DiDeMo (test) under $\Phi_\mathrm{vp}$, $\threshold=0.5$.
  The partly relevant ones include either ``{\it climb}'' or ``{\it kitty}'' in the query ``\textit{kitty first starts to climb}''.}
  \label{fig:didemo_qual_result1}
  \end{center}
\end{figure*}

\begin{figure*}
  \begin{center}    
   \includegraphics[width=0.9\linewidth]{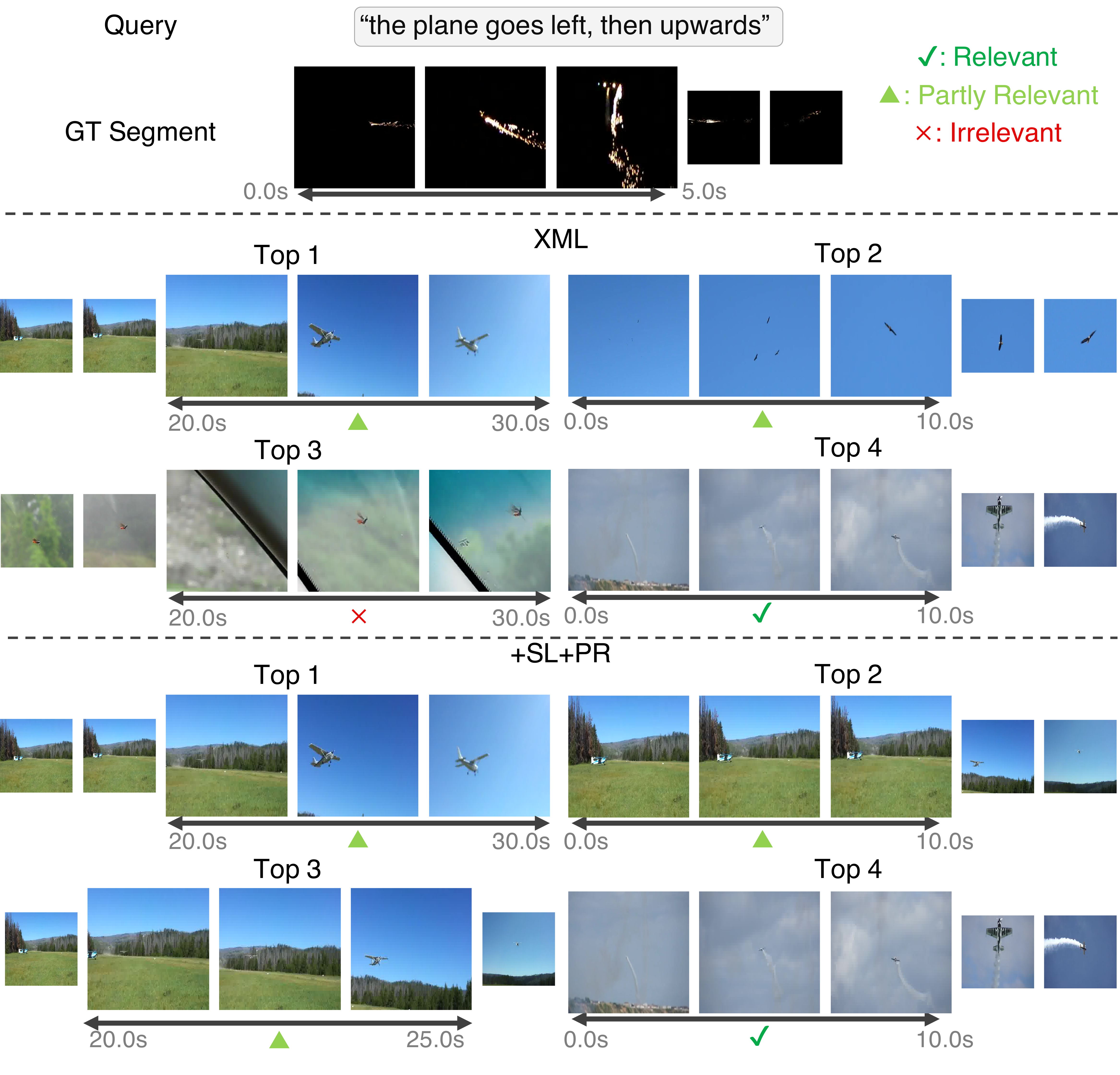}        
  \caption{Example of the qualitative results on DiDeMo (test) under $\Phi_\mathrm{vp}$, $\threshold=0.5$.
    The partly relevant ones include the scene of ``{\it plane}'' flying, 
    whereas the relevant one include ``\textit{goes left, then upwards}'' as well.}
  \label{fig:didemo_qual_result2}
  \end{center}
\end{figure*}

\begin{figure*}
  \begin{center}    
   \includegraphics[width=0.9\linewidth]{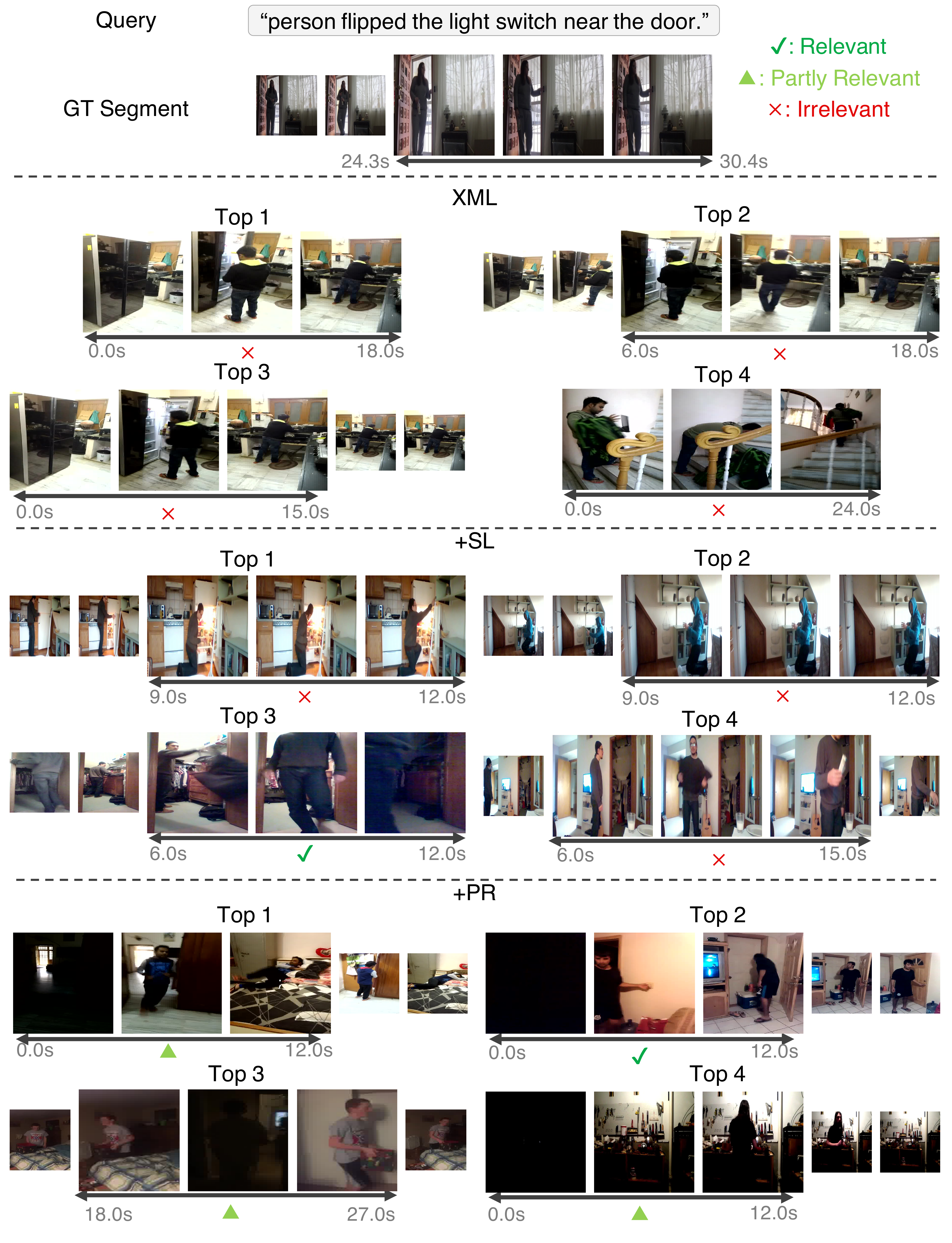}        
  \caption{Example of the qualitative results on Charades-STA (test) under $\Phi_\mathrm{np-vp}$, $\threshold=0.7$.
  The relevant ones certainly include ``\textit{person flipped the light switch near the door}'',
  whereas the partly relevant ones include the scene of light turning on/off even though it is uncertain if the ``\textit{person flipped the light switch}'' for sure.
  }
  \label{fig:charades_sta_qual_result1}
  \end{center}
\end{figure*}

\begin{figure*}
  \begin{center}    
   \includegraphics[width=0.9\linewidth]{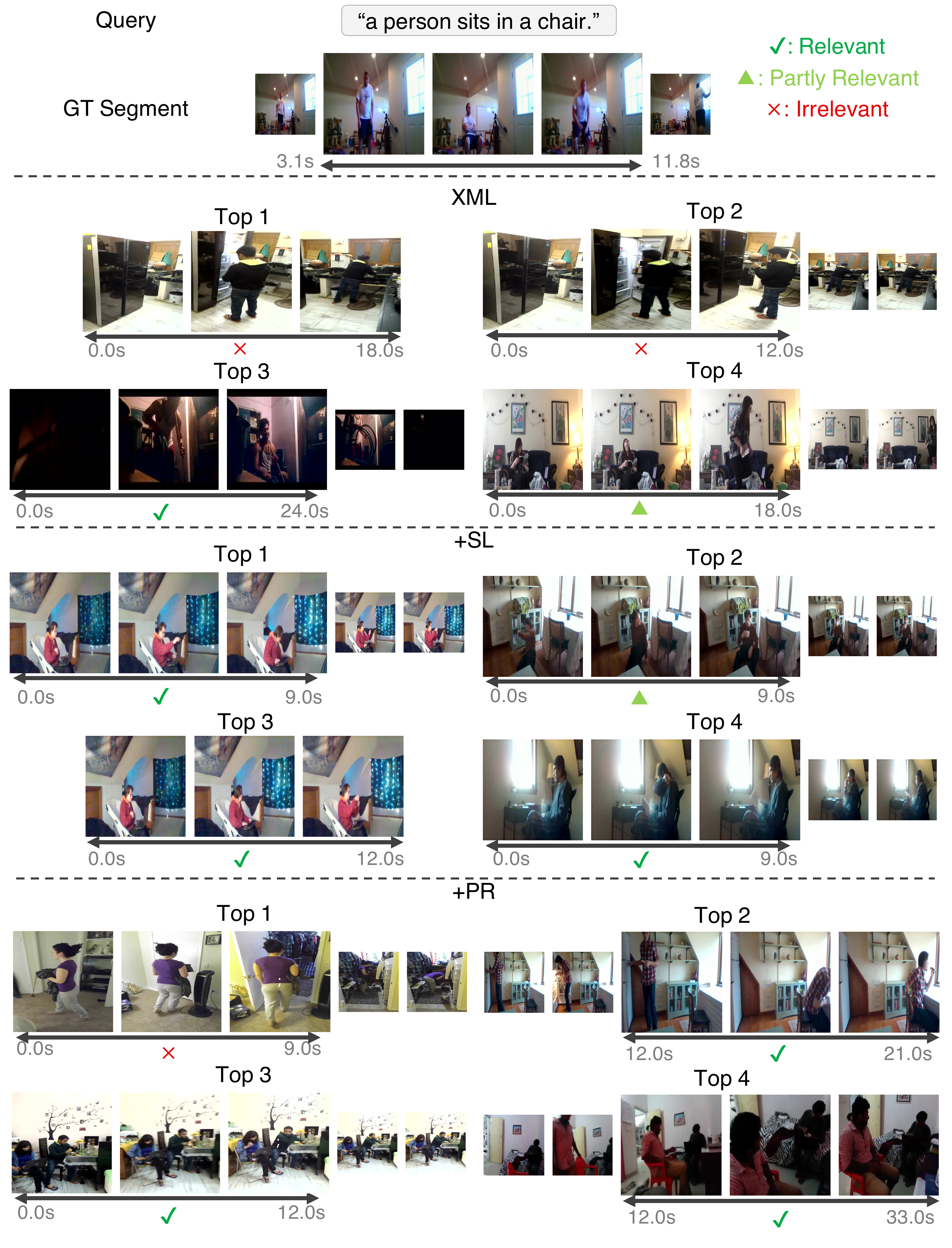}        
  \caption{Example of the qualitative results on Charades-STA (test) under $\Phi_\mathrm{np-vp}$, $\threshold=0.7$.
    The partly relevant ones include ``\textit{person sits}'' regardless of where they sit,
    whereas in the relevant ones ``\textit{person sits in a chair}''.}
  \label{fig:charades_sta_qual_result2}
  \end{center}
\end{figure*}

\begin{figure*}
  \begin{center}    
   \includegraphics[width=0.9\linewidth]{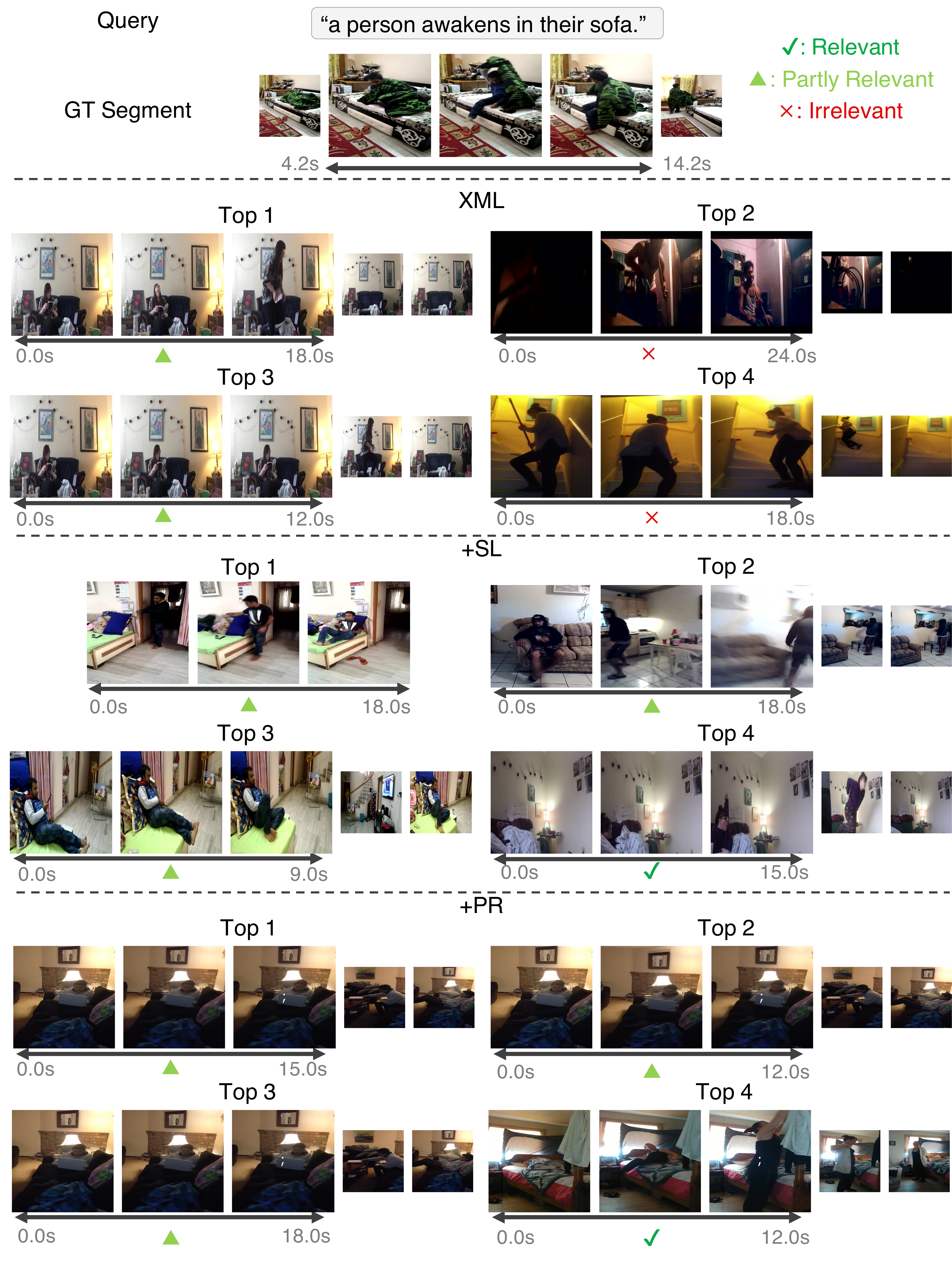}        
  \caption{Example of the qualitative results on Charades-STA (test) under $\Phi_\mathrm{np-vp}$, $\threshold=0.7$.
    The partly relevant ones include ``{\it sofa}'', whereas the relevant ones include ``\textit{a person awakens in their sofa}''.}
  \label{fig:charades_sta_qual_result3}
  \end{center}
\end{figure*}

Here are additional qualitative results on top of Fig. \ref{fig:didemo_qual_result3}.
\figref{fig:didemo_qual_result1} and \figref{fig:didemo_qual_result2} are results on DiDeMo,
whereas \figref{fig:charades_sta_qual_result1}, \figref{fig:charades_sta_qual_result2}, and \figref{fig:charades_sta_qual_result3} are results on Charades-STA.

The same tendency as in Sec. \ref{sec:qualitative_results} can be found in \figref{fig:didemo_qual_result1} and \figref{fig:didemo_qual_result2} as well.
In \figref{fig:didemo_qual_result1}, XML candidate moments are involved with inconsistent animals, 
which should be ``\textit{cat}'' because the query includes ``\textit{kitty}''.
Indeed, +SL+PR outputs the cat-related moments including the ground-truth moment.
Likewise, in \figref{fig:didemo_qual_result2}, though XML Top 3 moment is about ``\textit{flies}'',
all of the candidate moments of +SL+PR are involved with the ``\textit{plane}''.

As for Charades-STA, in \figref{fig:charades_sta_qual_result1}, \figref{fig:charades_sta_qual_result2}, and \figref{fig:charades_sta_qual_result3},
the results of +SL and +PR are mostly relevant to the input query.
Surprisingly, results of +SL are not much worse than those of +PR.
On the contrary, XML tends to give specific videos regardless of the input query as can be observed 
through \figref{fig:charades_sta_qual_result1}, \figref{fig:charades_sta_qual_result2}, and \figref{fig:charades_sta_qual_result3}.
This may suggest the limitation of the evaluation metrics.
Furthermore, the results in \figref{fig:charades_sta_qual_result1} are notable.
Due to the roundabout expression of the input query, XML and +SL tend to output irrelevant moments.
However, the results of +PR in \figref{fig:charades_sta_qual_result1} are impressing,
because all of the candidate moments are related to ``\textit{turn on/off the light switch}'',
which is implied by ``\textit{flipped the light switch}''
This is also probably thanks to SBERT as it can softly capture the semantics of the sentences.
Consequently, different sentences with similar semantics are considered to linked via training.

\bibliographystyle{ACM-Reference-Format}
\bibliography{egbib}


\begin{thebibliography}{34}


\ifx \showCODEN    \undefined \def \showCODEN     #1{\unskip}     \fi
\ifx \showDOI      \undefined \def \showDOI       #1{#1}\fi
\ifx \showISBNx    \undefined \def \showISBNx     #1{\unskip}     \fi
\ifx \showISBNxiii \undefined \def \showISBNxiii  #1{\unskip}     \fi
\ifx \showISSN     \undefined \def \showISSN      #1{\unskip}     \fi
\ifx \showLCCN     \undefined \def \showLCCN      #1{\unskip}     \fi
\ifx \shownote     \undefined \def \shownote      #1{#1}          \fi
\ifx \showarticletitle \undefined \def \showarticletitle #1{#1}   \fi
\ifx \showURL      \undefined \def \showURL       {\relax}        \fi
\providecommand\bibfield[2]{#2}
\providecommand\bibinfo[2]{#2}
\providecommand\natexlab[1]{#1}
\providecommand\showeprint[2][]{arXiv:#2}

\bibitem[\protect\citeauthoryear{Chen, Chen, Ma, Jie, and Chua}{Chen
  et~al\mbox{.}}{2018}]%
        {chen2018temporally}
\bibfield{author}{\bibinfo{person}{Jingyuan Chen}, \bibinfo{person}{Xinpeng
  Chen}, \bibinfo{person}{Lin Ma}, \bibinfo{person}{Zequn Jie}, {and}
  \bibinfo{person}{Tat-Seng Chua}.} \bibinfo{year}{2018}\natexlab{}.
\newblock \showarticletitle{{Temporally Grounding Natural Sentence in Video}}.
  In \bibinfo{booktitle}{\emph{EMNLP}}.
\newblock


\bibitem[\protect\citeauthoryear{Dembczynski, Cheng, and
  H^^c3^^bcllermeier}{Dembczynski et~al\mbox{.}}{2010}]%
        {DBLP:conf/icml/DembczynskiCH10}
\bibfield{author}{\bibinfo{person}{Krzysztof Dembczynski},
  \bibinfo{person}{Weiwei Cheng}, {and} \bibinfo{person}{Eyke
  H^^c3^^bcllermeier}.} \bibinfo{year}{2010}\natexlab{}.
\newblock \showarticletitle{{Bayes Optimal Multilabel Classification via
  Probabilistic Classifier Chains}}. In \bibinfo{booktitle}{\emph{ICML}}.
\newblock


\bibitem[\protect\citeauthoryear{Devlin, Chang, Lee, and Toutanova}{Devlin
  et~al\mbox{.}}{2019}]%
        {devlin-etal-2019-bert}
\bibfield{author}{\bibinfo{person}{Jacob Devlin}, \bibinfo{person}{Ming-Wei
  Chang}, \bibinfo{person}{Kenton Lee}, {and} \bibinfo{person}{Kristina
  Toutanova}.} \bibinfo{year}{2019}\natexlab{}.
\newblock \showarticletitle{{{BERT}: Pre-training of Deep Bidirectional
  Transformers for Language Understanding}}. In
  \bibinfo{booktitle}{\emph{ACL}}.
\newblock


\bibitem[\protect\citeauthoryear{Escorcia, Soldan, Sivic, Ghanem, and
  Russell}{Escorcia et~al\mbox{.}}{2019}]%
        {escorcia2019temporal}
\bibfield{author}{\bibinfo{person}{Victor Escorcia}, \bibinfo{person}{Mattia
  Soldan}, \bibinfo{person}{Josef Sivic}, \bibinfo{person}{Bernard Ghanem},
  {and} \bibinfo{person}{Bryan Russell}.} \bibinfo{year}{2019}\natexlab{}.
\newblock \showarticletitle{{Temporal Localization of Moments in Video
  Collections with Natural Language}}.
\newblock \bibinfo{journal}{\emph{arXiv preprint arXiv:1907.12763}}
  (\bibinfo{year}{2019}).
\newblock


\bibitem[\protect\citeauthoryear{Faghri, Fleet, Kiros, and Fidler}{Faghri
  et~al\mbox{.}}{2018}]%
        {faghri2018vse++}
\bibfield{author}{\bibinfo{person}{Fartash Faghri}, \bibinfo{person}{David~J
  Fleet}, \bibinfo{person}{Jamie~Ryan Kiros}, {and} \bibinfo{person}{Sanja
  Fidler}.} \bibinfo{year}{2018}\natexlab{}.
\newblock \showarticletitle{{VSE++: Improving Visual-Semantic Embeddings with
  Hard Negatives}}. In \bibinfo{booktitle}{\emph{BMVC}}.
\newblock
\urldef\tempurl%
\url{https://github.com/fartashf/vsepp}
\showURL{%
\tempurl}


\bibitem[\protect\citeauthoryear{Gabeur, Sun, Alahari, and Schmid}{Gabeur
  et~al\mbox{.}}{2020}]%
        {gabeur2020multi}
\bibfield{author}{\bibinfo{person}{Valentin Gabeur}, \bibinfo{person}{Chen
  Sun}, \bibinfo{person}{Karteek Alahari}, {and} \bibinfo{person}{Cordelia
  Schmid}.} \bibinfo{year}{2020}\natexlab{}.
\newblock \showarticletitle{{Multi-modal Transformer for Video Retrieval}}. In
  \bibinfo{booktitle}{\emph{ECCV}}.
\newblock


\bibitem[\protect\citeauthoryear{Gao, Sun, Yang, and Nevatia}{Gao
  et~al\mbox{.}}{2017}]%
        {gao2017tall}
\bibfield{author}{\bibinfo{person}{Jiyang Gao}, \bibinfo{person}{Chen Sun},
  \bibinfo{person}{Zhenheng Yang}, {and} \bibinfo{person}{Ram Nevatia}.}
  \bibinfo{year}{2017}\natexlab{}.
\newblock \showarticletitle{{TALL: Temporal Activity Localization via Language
  Query}}. In \bibinfo{booktitle}{\emph{ICCV}}.
\newblock


\bibitem[\protect\citeauthoryear{He, Zhang, Ren, and Sun}{He
  et~al\mbox{.}}{2016}]%
        {he2016deep}
\bibfield{author}{\bibinfo{person}{Kaiming He}, \bibinfo{person}{Xiangyu
  Zhang}, \bibinfo{person}{Shaoqing Ren}, {and} \bibinfo{person}{Jian Sun}.}
  \bibinfo{year}{2016}\natexlab{}.
\newblock \showarticletitle{{Deep Residual Learning for Image Recognition}}. In
  \bibinfo{booktitle}{\emph{CVPR}}.
\newblock


\bibitem[\protect\citeauthoryear{Hendricks, Wang, Shechtman, Sivic, Darrell,
  and Russell}{Hendricks et~al\mbox{.}}{2017}]%
        {hendricks17iccv}
\bibfield{author}{\bibinfo{person}{Lisa~Anne Hendricks},
  \bibinfo{person}{Oliver Wang}, \bibinfo{person}{Eli Shechtman},
  \bibinfo{person}{Josef Sivic}, \bibinfo{person}{Trevor Darrell}, {and}
  \bibinfo{person}{Bryan Russell}.} \bibinfo{year}{2017}\natexlab{}.
\newblock \showarticletitle{{Localizing Moments in Video with Natural
  Language}}. In \bibinfo{booktitle}{\emph{ICCV}}.
\newblock


\bibitem[\protect\citeauthoryear{Hendricks, Wang, Shechtman, Sivic, Darrell,
  and Russell}{Hendricks et~al\mbox{.}}{2018}]%
        {hendricks18emnlp}
\bibfield{author}{\bibinfo{person}{Lisa~Anne Hendricks},
  \bibinfo{person}{Oliver Wang}, \bibinfo{person}{Eli Shechtman},
  \bibinfo{person}{Josef Sivic}, \bibinfo{person}{Trevor Darrell}, {and}
  \bibinfo{person}{Bryan Russell}.} \bibinfo{year}{2018}\natexlab{}.
\newblock \showarticletitle{{Localizing Moments in Video with Temporal
  Language.}}. In \bibinfo{booktitle}{\emph{EMNLP}}.
\newblock


\bibitem[\protect\citeauthoryear{Jaskie and Spanias}{Jaskie and
  Spanias}{2019}]%
        {jaskie2019positive}
\bibfield{author}{\bibinfo{person}{Kristen Jaskie} {and}
  \bibinfo{person}{Andreas Spanias}.} \bibinfo{year}{2019}\natexlab{}.
\newblock \showarticletitle{{POSITIVE AND UNLABELED LEARNING ALGORITHMS AND
  APPLICATIONS: A SURVEY }}. In \bibinfo{booktitle}{\emph{IISA}}.
\newblock


\bibitem[\protect\citeauthoryear{Kanehira and Harada}{Kanehira and
  Harada}{2016}]%
        {kanehira2016multi}
\bibfield{author}{\bibinfo{person}{Atsushi Kanehira} {and}
  \bibinfo{person}{Tatsuya Harada}.} \bibinfo{year}{2016}\natexlab{}.
\newblock \showarticletitle{{Multi-label Ranking from Positive and Unlabeled
  Data}}. In \bibinfo{booktitle}{\emph{CVPR}}.
\newblock


\bibitem[\protect\citeauthoryear{Kingma and Ba}{Kingma and Ba}{2015}]%
        {kingma:adam}
\bibfield{author}{\bibinfo{person}{Diederick~P Kingma} {and}
  \bibinfo{person}{Jimmy Ba}.} \bibinfo{year}{2015}\natexlab{}.
\newblock \showarticletitle{{Adam: A Method for Stochastic Optimization}}. In
  \bibinfo{booktitle}{\emph{ICLR}}.
\newblock


\bibitem[\protect\citeauthoryear{Kitaev and Klein}{Kitaev and Klein}{2018}]%
        {Kitaev-2018-SelfAttentive}
\bibfield{author}{\bibinfo{person}{Nikita Kitaev} {and} \bibinfo{person}{Dan
  Klein}.} \bibinfo{year}{2018}\natexlab{}.
\newblock \showarticletitle{{Constituency Parsing with a Self-Attentive
  Encoder}}. In \bibinfo{booktitle}{\emph{ACL}}.
\newblock


\bibitem[\protect\citeauthoryear{Lei, Yu, Berg, and Bansal}{Lei
  et~al\mbox{.}}{2020}]%
        {lei2020tvr}
\bibfield{author}{\bibinfo{person}{Jie Lei}, \bibinfo{person}{Licheng Yu},
  \bibinfo{person}{Tamara~L Berg}, {and} \bibinfo{person}{Mohit Bansal}.}
  \bibinfo{year}{2020}\natexlab{}.
\newblock \showarticletitle{{TVR: A Large-Scale Dataset for Video-Subtitle
  Moment Retrieval}}. In \bibinfo{booktitle}{\emph{ECCV}}.
\newblock


\bibitem[\protect\citeauthoryear{Li, Chen, Cheng, Gan, Yu, and Liu}{Li
  et~al\mbox{.}}{2020}]%
        {li2020hero}
\bibfield{author}{\bibinfo{person}{Linjie Li}, \bibinfo{person}{Yen-Chun Chen},
  \bibinfo{person}{Yu Cheng}, \bibinfo{person}{Zhe Gan},
  \bibinfo{person}{Licheng Yu}, {and} \bibinfo{person}{Jingjing Liu}.}
  \bibinfo{year}{2020}\natexlab{}.
\newblock \showarticletitle{{HERO: Hierarchical Encoder for Video+ Language
  Omni-representation Pre-training}}. In \bibinfo{booktitle}{\emph{EMNLP}}.
\newblock


\bibitem[\protect\citeauthoryear{Lin, Cai, Wang, Liu, Zheng, and Shi}{Lin
  et~al\mbox{.}}{2020}]%
        {lin2020world}
\bibfield{author}{\bibinfo{person}{Zibo Lin}, \bibinfo{person}{Deng Cai},
  \bibinfo{person}{Yan Wang}, \bibinfo{person}{Xiaojiang Liu},
  \bibinfo{person}{Haitao Zheng}, {and} \bibinfo{person}{Shuming Shi}.}
  \bibinfo{year}{2020}\natexlab{}.
\newblock \showarticletitle{{The World Is Not Binary: Learning to Rank with
  Grayscale Data for Dialogue Response Selection}}. In
  \bibinfo{booktitle}{\emph{EMNLP}}.
\newblock


\bibitem[\protect\citeauthoryear{Liu, Yeung, Chou, Huang, Fei-Fei, and
  Niebles}{Liu et~al\mbox{.}}{2018}]%
        {liu2018tmn}
\bibfield{author}{\bibinfo{person}{Bingbin Liu}, \bibinfo{person}{Serena
  Yeung}, \bibinfo{person}{Edward Chou}, \bibinfo{person}{De-An Huang},
  \bibinfo{person}{Li Fei-Fei}, {and} \bibinfo{person}{Juan~Carlos Niebles}.}
  \bibinfo{year}{2018}\natexlab{}.
\newblock \showarticletitle{{Temporal Modular Networks for Retrieving Complex
  Compositional Activities in Videos}}. In \bibinfo{booktitle}{\emph{ECCV}}.
\newblock


\bibitem[\protect\citeauthoryear{Liu, Ott, Goyal, Du, Joshi, Chen, Levy, Lewis,
  Zettlemoyer, and Stoyanov}{Liu et~al\mbox{.}}{2019}]%
        {liu2019roberta}
\bibfield{author}{\bibinfo{person}{Yinhan Liu}, \bibinfo{person}{Myle Ott},
  \bibinfo{person}{Naman Goyal}, \bibinfo{person}{Jingfei Du},
  \bibinfo{person}{Mandar Joshi}, \bibinfo{person}{Danqi Chen},
  \bibinfo{person}{Omer Levy}, \bibinfo{person}{Mike Lewis},
  \bibinfo{person}{Luke Zettlemoyer}, {and} \bibinfo{person}{Veselin
  Stoyanov}.} \bibinfo{year}{2019}\natexlab{}.
\newblock \showarticletitle{{RoBERTa: A Robustly Optimized BERT Pretraining
  Approach}}.
\newblock \bibinfo{journal}{\emph{arXiv preprint arXiv:1907.11692}}
  (\bibinfo{year}{2019}).
\newblock


\bibitem[\protect\citeauthoryear{Maeoki, Uehara, and Harada}{Maeoki
  et~al\mbox{.}}{2020}]%
        {Maeoki_2020_CVPR_Workshops}
\bibfield{author}{\bibinfo{person}{Sho Maeoki}, \bibinfo{person}{Kohei Uehara},
  {and} \bibinfo{person}{Tatsuya Harada}.} \bibinfo{year}{2020}\natexlab{}.
\newblock \showarticletitle{{Interactive Video Retrieval with Dialog}}. In
  \bibinfo{booktitle}{\emph{CVPR Workshop}}.
\newblock


\bibitem[\protect\citeauthoryear{Miech, Laptev, and Sivic}{Miech
  et~al\mbox{.}}{2018}]%
        {miech18learning}
\bibfield{author}{\bibinfo{person}{Antoine Miech}, \bibinfo{person}{Ivan
  Laptev}, {and} \bibinfo{person}{Josef Sivic}.}
  \bibinfo{year}{2018}\natexlab{}.
\newblock \showarticletitle{Learning a {T}ext-{V}ideo {E}mbedding from
  {I}ncomplete and {H}eterogeneous {D}ata}.
\newblock \bibinfo{journal}{\emph{arXiv preprint arXiv:1804.02516}}
  (\bibinfo{year}{2018}).
\newblock


\bibitem[\protect\citeauthoryear{Mithun, Li, Metze, and Roy-Chowdhury}{Mithun
  et~al\mbox{.}}{2018}]%
        {mithun2018learning}
\bibfield{author}{\bibinfo{person}{Niluthpol~Chowdhury Mithun},
  \bibinfo{person}{Juncheng Li}, \bibinfo{person}{Florian Metze}, {and}
  \bibinfo{person}{Amit~K Roy-Chowdhury}.} \bibinfo{year}{2018}\natexlab{}.
\newblock \showarticletitle{{Learning Joint Embedding with Multimodal Cues for
  Cross-Modal Video-Text Retrieval}}. In \bibinfo{booktitle}{\emph{ICMR}}.
\newblock


\bibitem[\protect\citeauthoryear{Papineni, Roukos, Ward, and Zhu}{Papineni
  et~al\mbox{.}}{2002}]%
        {papineni-etal-2002-bleu}
\bibfield{author}{\bibinfo{person}{Kishore Papineni}, \bibinfo{person}{Salim
  Roukos}, \bibinfo{person}{Todd Ward}, {and} \bibinfo{person}{Wei-Jing Zhu}.}
  \bibinfo{year}{2002}\natexlab{}.
\newblock \showarticletitle{{BLEU: a Method for Automatic Evaluation of Machine
  Translation}}. In \bibinfo{booktitle}{\emph{ACL}}.
\newblock


\bibitem[\protect\citeauthoryear{Paszke, Gross, Massa, Lerer, Bradbury, Chanan,
  Killeen, Lin, Gimelshein, Antiga, Desmaison, Kopf, Yang, DeVito, Raison,
  Tejani, Chilamkurthy, Steiner, Fang, Bai, and Chintala}{Paszke
  et~al\mbox{.}}{2019}]%
        {NEURIPS2019_9015}
\bibfield{author}{\bibinfo{person}{Adam Paszke}, \bibinfo{person}{Sam Gross},
  \bibinfo{person}{Francisco Massa}, \bibinfo{person}{Adam Lerer},
  \bibinfo{person}{James Bradbury}, \bibinfo{person}{Gregory Chanan},
  \bibinfo{person}{Trevor Killeen}, \bibinfo{person}{Zeming Lin},
  \bibinfo{person}{Natalia Gimelshein}, \bibinfo{person}{Luca Antiga},
  \bibinfo{person}{Alban Desmaison}, \bibinfo{person}{Andreas Kopf},
  \bibinfo{person}{Edward Yang}, \bibinfo{person}{Zachary DeVito},
  \bibinfo{person}{Martin Raison}, \bibinfo{person}{Alykhan Tejani},
  \bibinfo{person}{Sasank Chilamkurthy}, \bibinfo{person}{Benoit Steiner},
  \bibinfo{person}{Lu Fang}, \bibinfo{person}{Junjie Bai}, {and}
  \bibinfo{person}{Soumith Chintala}.} \bibinfo{year}{2019}\natexlab{}.
\newblock \showarticletitle{{PyTorch: An Imperative Style, High-Performance
  Deep Learning Library}}. In \bibinfo{booktitle}{\emph{NeurIPS}}.
\newblock


\bibitem[\protect\citeauthoryear{Paul, Mithun, and Roy{-}Chowdhury}{Paul
  et~al\mbox{.}}{2020}]%
        {sudipta2020text}
\bibfield{author}{\bibinfo{person}{Sudipta Paul},
  \bibinfo{person}{Niluthpol~Chowdhury Mithun}, {and} \bibinfo{person}{Amit~K.
  Roy{-}Chowdhury}.} \bibinfo{year}{2020}\natexlab{}.
\newblock \showarticletitle{{Text-based Localization of Moments in a Video
  Corpus}}.
\newblock \bibinfo{journal}{\emph{arXiv preprint arXiv:2008.08716}}
  (\bibinfo{year}{2020}).
\newblock


\bibitem[\protect\citeauthoryear{Reimers and Gurevych}{Reimers and
  Gurevych}{2019}]%
        {reimers-gurevych-2019-sentence}
\bibfield{author}{\bibinfo{person}{Nils Reimers} {and} \bibinfo{person}{Iryna
  Gurevych}.} \bibinfo{year}{2019}\natexlab{}.
\newblock \showarticletitle{{Sentence-{BERT}: Sentence Embeddings using
  {S}iamese {BERT}-Networks}}. In \bibinfo{booktitle}{\emph{EMNLP-IJCNLP}}.
\newblock


\bibitem[\protect\citeauthoryear{Sanh, Debut, Chaumond, and Wolf}{Sanh
  et~al\mbox{.}}{2019}]%
        {sanh2019distilbert}
\bibfield{author}{\bibinfo{person}{Victor Sanh}, \bibinfo{person}{Lysandre
  Debut}, \bibinfo{person}{Julien Chaumond}, {and} \bibinfo{person}{Thomas
  Wolf}.} \bibinfo{year}{2019}\natexlab{}.
\newblock \showarticletitle{{DistilBERT, a distilled version of BERT: smaller,
  faster, cheaper and lighter}}. In \bibinfo{booktitle}{\emph{NeurIPS
  Workshop}}.
\newblock


\bibitem[\protect\citeauthoryear{Sigurdsson, Varol, Wang, Farhadi, Laptev, and
  Gupta}{Sigurdsson et~al\mbox{.}}{2016}]%
        {sigurdsson2016hollywood}
\bibfield{author}{\bibinfo{person}{Gunnar~A Sigurdsson},
  \bibinfo{person}{G{\"u}l Varol}, \bibinfo{person}{Xiaolong Wang},
  \bibinfo{person}{Ali Farhadi}, \bibinfo{person}{Ivan Laptev}, {and}
  \bibinfo{person}{Abhinav Gupta}.} \bibinfo{year}{2016}\natexlab{}.
\newblock \showarticletitle{{Hollywood in Homes: Crowdsourcing Data Collection
  for Activity Understanding}}. In \bibinfo{booktitle}{\emph{ECCV}}.
\newblock


\bibitem[\protect\citeauthoryear{Vaswani, Shazeer, Parmar, Uszkoreit, Jones,
  Gomez, Kaiser, and Polosukhin}{Vaswani et~al\mbox{.}}{2017}]%
        {vaswani2017attention}
\bibfield{author}{\bibinfo{person}{Ashish Vaswani}, \bibinfo{person}{Noam
  Shazeer}, \bibinfo{person}{Niki Parmar}, \bibinfo{person}{Jakob Uszkoreit},
  \bibinfo{person}{Llion Jones}, \bibinfo{person}{Aidan~N Gomez},
  \bibinfo{person}{{\L}ukasz Kaiser}, {and} \bibinfo{person}{Illia
  Polosukhin}.} \bibinfo{year}{2017}\natexlab{}.
\newblock \showarticletitle{{Attention is All You Need}}. In
  \bibinfo{booktitle}{\emph{NeurIPS}}.
\newblock


\bibitem[\protect\citeauthoryear{Young, Lai, Hodosh, and Hockenmaier}{Young
  et~al\mbox{.}}{2014}]%
        {young-etal-2014-image}
\bibfield{author}{\bibinfo{person}{Peter Young}, \bibinfo{person}{Alice Lai},
  \bibinfo{person}{Micah Hodosh}, {and} \bibinfo{person}{Julia Hockenmaier}.}
  \bibinfo{year}{2014}\natexlab{}.
\newblock \showarticletitle{{From image descriptions to visual denotations: New
  similarity metrics for semantic inference over event descriptions}}.
\newblock \bibinfo{journal}{\emph{Transactions of the Association for
  Computational Linguistics}}  \bibinfo{volume}{2} (\bibinfo{year}{2014}),
  \bibinfo{pages}{67--78}.
\newblock


\bibitem[\protect\citeauthoryear{Yu, Kim, and Kim}{Yu et~al\mbox{.}}{2018}]%
        {yu2018joint}
\bibfield{author}{\bibinfo{person}{Youngjae Yu}, \bibinfo{person}{Jongseok
  Kim}, {and} \bibinfo{person}{Gunhee Kim}.} \bibinfo{year}{2018}\natexlab{}.
\newblock \showarticletitle{{A Joint Sequence Fusion Model for Video Question
  Answering and Retrieval}}. In \bibinfo{booktitle}{\emph{ECCV}}.
\newblock


\bibitem[\protect\citeauthoryear{Zhang, Hu, Jain, Ie, and Sha}{Zhang
  et~al\mbox{.}}{2020a}]%
        {zhang-etal-2020-learning}
\bibfield{author}{\bibinfo{person}{Bowen Zhang}, \bibinfo{person}{Hexiang Hu},
  \bibinfo{person}{Vihan Jain}, \bibinfo{person}{Eugene Ie}, {and}
  \bibinfo{person}{Fei Sha}.} \bibinfo{year}{2020}\natexlab{a}.
\newblock \showarticletitle{{Learning to Represent Image and Text with
  Denotation Graph}}. In \bibinfo{booktitle}{\emph{EMNLP}}.
\newblock


\bibitem[\protect\citeauthoryear{Zhang, Hu, Lee, Zhao, Chammas, Jain, Ie, and
  Sha}{Zhang et~al\mbox{.}}{2020b}]%
        {zhang2020hierarchical}
\bibfield{author}{\bibinfo{person}{Bowen Zhang}, \bibinfo{person}{Hexiang Hu},
  \bibinfo{person}{Joonseok Lee}, \bibinfo{person}{Ming Zhao},
  \bibinfo{person}{Sheide Chammas}, \bibinfo{person}{Vihan Jain},
  \bibinfo{person}{Eugene Ie}, {and} \bibinfo{person}{Fei Sha}.}
  \bibinfo{year}{2020}\natexlab{b}.
\newblock \showarticletitle{{A Hierarchical Multi-Modal Encoder for Moment
  Localization in Video Corpus}}.
\newblock \bibinfo{journal}{\emph{arXiv preprint arXiv:2011.09046}}
  (\bibinfo{year}{2020}).
\newblock


\bibitem[\protect\citeauthoryear{Zhang, Dai, Wang, Wang, and Davis}{Zhang
  et~al\mbox{.}}{2019}]%
        {zhang2019man}
\bibfield{author}{\bibinfo{person}{Da Zhang}, \bibinfo{person}{Xiyang Dai},
  \bibinfo{person}{Xin Wang}, \bibinfo{person}{Yuan-Fang Wang}, {and}
  \bibinfo{person}{Larry~S Davis}.} \bibinfo{year}{2019}\natexlab{}.
\newblock \showarticletitle{{MAN: Moment Alignment Network for Natural Language
  Moment Retrieval via Iterative Graph Adjustment}}. In
  \bibinfo{booktitle}{\emph{CVPR}}.
\newblock


\end{thebibliography}

\end{document}